\documentclass[11pt]{article}

\usepackage[final]{acl}

\usepackage{times}
\usepackage{latexsym}

\usepackage[T1]{fontenc}

\usepackage[utf8]{inputenc}

\usepackage{microtype}

\usepackage{inconsolata}

\usepackage{graphicx}

\usepackage{enumitem}
\usepackage{multirow}
\usepackage{amssymb}

\usepackage[table]{xcolor}
\usepackage{booktabs}       
\usepackage{tabularx}
\usepackage{cuted}
\usepackage{amsmath}
\usepackage{xcolor}

\definecolor{thinkblue}{RGB}{255, 134, 58}
\definecolor{toolgreen}{RGB}{0, 176, 240}
\definecolor{resp-purple}{RGB}{112, 173, 72}
\definecolor{lightpurple}{RGB}{230,230,250} 

\title{\textsc{ToolCAD}: Exploring Tool-Using Large Language Models in Text-to-CAD Generation with Reinforcement Learning}

\author{
  \textbf{Yifei Gong\textsuperscript{1}},
  \textbf{Xing Wu\textsuperscript{1*}},
  \textbf{Wenda Liu\textsuperscript{1}},
  \textbf{Kang Tu\textsuperscript{1}},
\\
  \textsuperscript{1}School of Computer Engineering \& Science, Shanghai University
\\
  \small{
    \textbf{Correspondence:} \href{mailto:xingwu@shu.edu.cn}{xingwu@shu.edu.cn}
  }
}

\usepackage{cuted}
\usepackage{tcolorbox} 
\tcbuselibrary{skins, breakable}
\tcbset{
  myboxstyle/.style={
    colback=white,       
    colframe=black,      
    coltitle=white,      
    colbacktitle=black,  
    boxrule=0.5mm,      
    arc=2.5mm,             
    width=\linewidth,    
    boxsep=5pt,          
    breakable,            
    halign title=center
  }
}

\usepackage{listings}
\usepackage{courier} 
\usepackage[ruled,vlined]{algorithm2e}
\definecolor{bggray}{rgb}{0.95,0.95,0.95}
\definecolor{docstring}{rgb}{0.72,0.17,0.14}
\definecolor{keywordblue}{rgb}{0.13,0.13,1}

\lstdefinestyle{pythoncode}{
  backgroundcolor=\color{bggray},      
  language=Python,
  basicstyle=\ttfamily\small,
  keywordstyle=\color{keywordblue},
  stringstyle=\color{docstring},
  commentstyle=\itshape\color{gray},
  showstringspaces=false,
  breaklines=true,
  frame=single,
  rulecolor=\color{gray},
  numbers=left,
  numberstyle=\tiny\color{gray},
  tabsize=4,
  captionpos=b
}

\begin{document}
\maketitle
\begin{abstract}
Computer-Aided Design (CAD) is an expert-level task that relies on long-horizon reasoning and coherent modeling actions. Large Language Models (LLMs) have shown remarkable advancements in enabling language agents to tackle real-world tasks. Notably, there has been no investigation into how tool-using LLMs optimally interact with CAD engines, hindering the emergence of LLM-based agentic text-to-CAD modeling systems. We propose \textsc{ToolCAD}, a novel agentic CAD framework deploying LLMs as tool-using agents for text-to-CAD generation. Furthermore, we introduce an interactive CAD modeling gym to rollout reasoning and tool-augmented interaction trajectories with the CAD engine, incorporating hybrid feedback and human supervision. Meanwhile, an end-to-end post-training strategy is presented to enable LLMs to elicit refined CAD Modeling Chain of Thought (CAD-CoT) and evolve into proficient CAD tool-using agents via curriculum online reinforcement learning. Our findings demonstrate \textsc{ToolCAD} fills the gap in adopting and training open-source LLMs for CAD tool-using agents, enabling them to perform comparably to proprietary models, paving the way for more accessible and robust autonomous text-to-CAD modeling systems.\footnote{We make our project available on \url{https://gongyifeiisme.github.io/toolcad-project}.}
\end{abstract}

\section{Introduction}
\label{sec:intro}
Computer-Aided Design (CAD) serves as a fundamental tool across the entire product lifecycle in modern industrial manufacturing, supporting continuous tracking and iterative refinement \cite{cherng1998feature}. However, prototyping for industrial production from scratch still requires expert designers to manually perform accurate modeling with geometry-aware understanding, which is time-consuming and hinders the development of automation in complex CAD modeling.
\begin{figure}[!t]
    \centering

    \includegraphics[width=\linewidth]{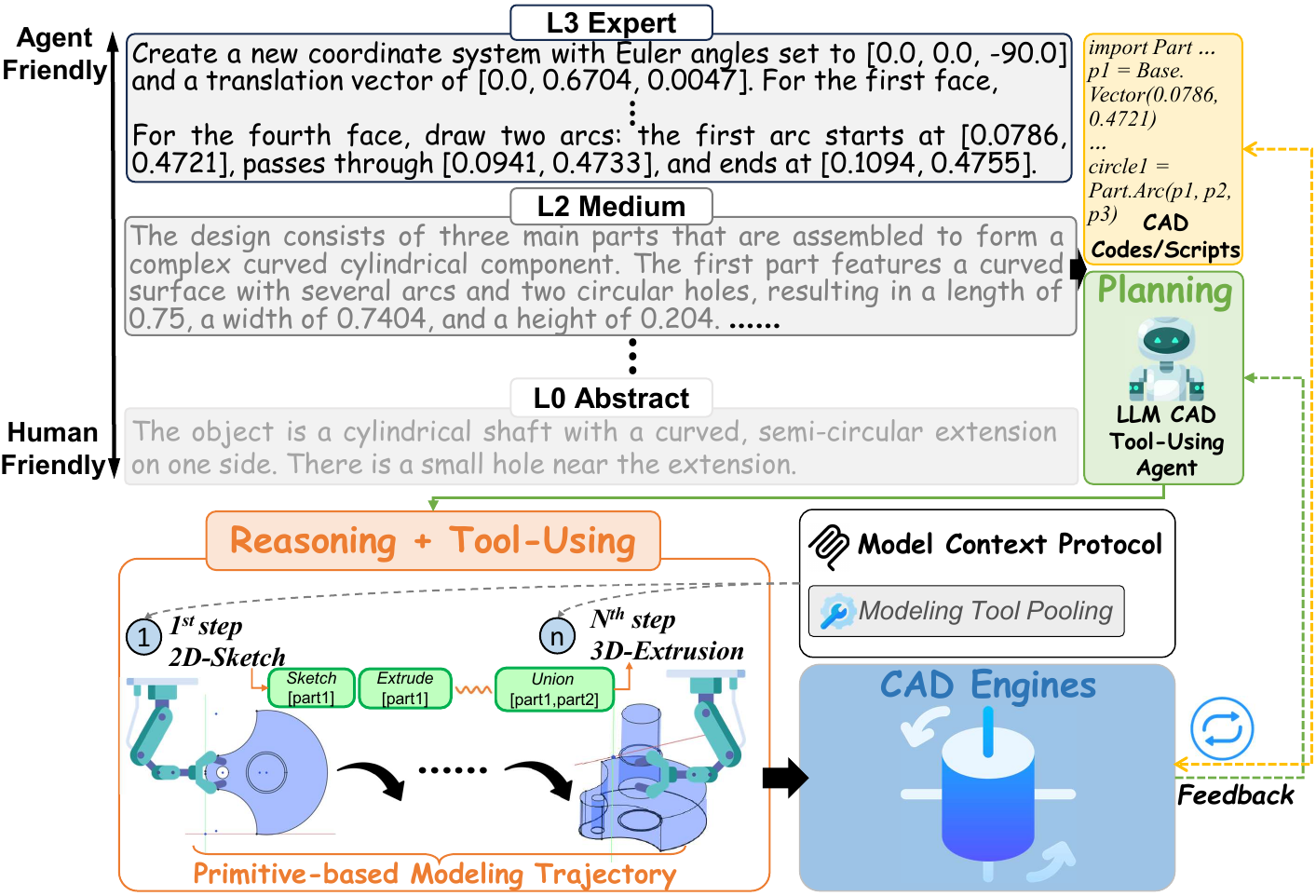}
    
    \captionof{figure}{\textbf{Prompt-to-Tool vs. Prompt-to-Code}. L3 expert-level modeling text enables CAD tool-using agents to plan, reason, call tools, and complete modeling tasks via iterative CAD-engine feedback.}
  
    \label{fig:idea}

\end{figure}
To enhance modeling efficiency and enable automation, modern CAD products  (\emph{e.g., FreeCAD, SolidWorks}) provide APIs with procedural scripting codes such as Python for rapid modeling \cite{badagabettu2024query2cad}. Besides, recent research has demonstrated that AI Agent for CAD generation, empowered by Large Language Models (LLMs) and Vision Language Models (VLMs), can increase designer productivity \cite{makatura2023can, kodnongbua2023zero}. Nevertheless, despite these advances, there remains a significant gap between existing methods and a fully autonomous expert CAD modeling system that seamlessly integrates with design platforms\cite{zhou2025status}. Also, it remains unclear how to enable LLMs to directly interact optimally with CAD engines, rather than relying on token-based predictive modeling or generating CAD codes\cite{text2cad,kolodiazhnyi2025cadrille}. Inspired by complex tool invocation in systems like Search-R1\cite{jin2025search} via LLM's external tool-use capabilities, leveraging long-horizon reasoning with tool-integrated learning, we aim to develop LLM-based CAD tool-using agents capable of automatic modeling guided by natural language commands (see Fig.\ref{fig:idea}). 

In contrast to agentic CAD generation methods \cite{mallis2024cad,zhang2024flexcad,wang2025cad}, our research explores the potential of LLMs leveraging reasoning and primitive-based tool usage for autonomous text-to-CAD generation. In this paper, the LLM acts as an expert tool-augmented CAD agent, tailored for decision-making in modeling steps and executing corresponding modeling tools \cite{xi2024agentgym, qian2025toolrl}. However, applying language agents to interact with the CAD engine and realize the full automation of CAD modeling workflow presents three key challenges: (1) Limited reasoning and tool-integration capabilities. (2) Lack of interactive CAD environments with modeling feedback, and agentic CAD tool-using evaluation benchmarks. (3) Insufficient training for proficient CAD tool-using agents across complex text-to-CAD tasks.

To address the aforementioned challenges, we propose \textsc{ToolCAD}, an agentic CAD framework, by leveraging the in-context Chain-of-Thought prompting techniques and further online reinforcement learning (RL) to unlock the LLMs' capabilities in step-level CAD Chain-of-Thought (CAD-CoT) reasoning and tool integration. This framework introduces an interactive CAD-specific modeling gym with hybrid feedback and human supervision, providing step-level and trajectory-level reward to enable LLMs to generate correct modeling tool-using trajectories through optimal interaction with the CAD engine. To further tackle CAD modeling tasks of varying complexity, we adopt a part-wise CAD curriculum exploration strategy and online GRPO optimization to improve the modeling stability of prompt-based agent policy, thereby developing robust and generalizable tool-using agents for complex text-to-CAD tasks. 

Our contributions can be summarized as follows:
\begin{itemize}[itemsep=1.8pt, topsep=0pt, parsep=0pt]
  \item We propose \textsc{ToolCAD}, a novel framework that achieves full automation of text-to-CAD generation by leveraging tool-using LLM agents. 
  \item \textsc{ToolCAD} advances RL for tool-using LLM end-to-end training by introducing interactive CAD modeling gym with hybrid supervision from rule-based and outcome-based feedback signals. It provides a curriculum-based online exploration tool-learning strategy to develop more competitive CAD tool-using agents.
  \item We demonstrate that \textsc{ToolCAD} outperforms agentic generation baselines, enabling high-quality text-to-CAD results on held-out tasks across diverse complexities.
\end{itemize}




\section{Related Work}
\label{sec:related_work}
\textbf{Intelligent CAD Modeling System.} Under the current trend of LLMs and VLMs demonstrating strong capabilities in complex reasoning and planning for real-world tasks, their integration into generative CAD modeling holds great potential for realizing intelligent and autonomous design systems. Recent advances on LLM-based CAD modeling approaches can be broadly categorized into two directions: \textbf{(1) LLM-based Parametric CAD Sequences Generation}: leveraging LLMs as auxiliary modules to assist token-level prediction for the next modeling command. Text2CAD \cite{text2cad} generates parametric CAD sequences from natural language instructions using a transformer-based network pretrained on multi-level CAD prompts annotated by Mistral and LLaVA-NeXT. CAD-GPT \cite{wang2025cad} leverages the Multimodal Large Language Model (MLLM) LLaVA-1.5-7B, enhanced with 3D spatial reasoning capabilities, to precisely synthesize CAD modeling sequences. CAD-MLLM \cite{xu2024cad}, the first intelligent CAD system to use a LLM Vicuna-7B to align multimodal input with modeling commands sequences for generating CAD models. \textbf{(2) LLM-based CAD Code Generation}: utilizing LLMs to generate and refine code-based modeling instructions for CAD reconstruction. CAD-Assistant \cite{mallis2024cad} uses GPT-4o with tool-augmented and docstring prompting to plan and generate code-level actions for autoconstraining and sketch parameterization. CAD-Llama \cite{li2025cad} augments CAD code generation using instruction-tuned LLaMA3-8B with Structured Parametric CAD Code (SPCC). CADCodeVerify \cite{alrashedy2024generating} prompts GPT-4o to make decisions on CAD design adjustments through an interactive question-answer feedback for code refinement. Seek-CAD \cite{li2025seek} integrates both visual and CoT feedback from DeepSeek-R1 and Gemini-2.0 to enable self-refinement of CAD code. Unlike these methods, \textsc{ToolCAD} explores the optimal strategy for primitive-based tool-using LLM agents to interact with CAD engines for text-to-CAD generation.
\label{sec:framework}
\begin{figure*}[ht]
    \centering
    \includegraphics[width=1\linewidth]{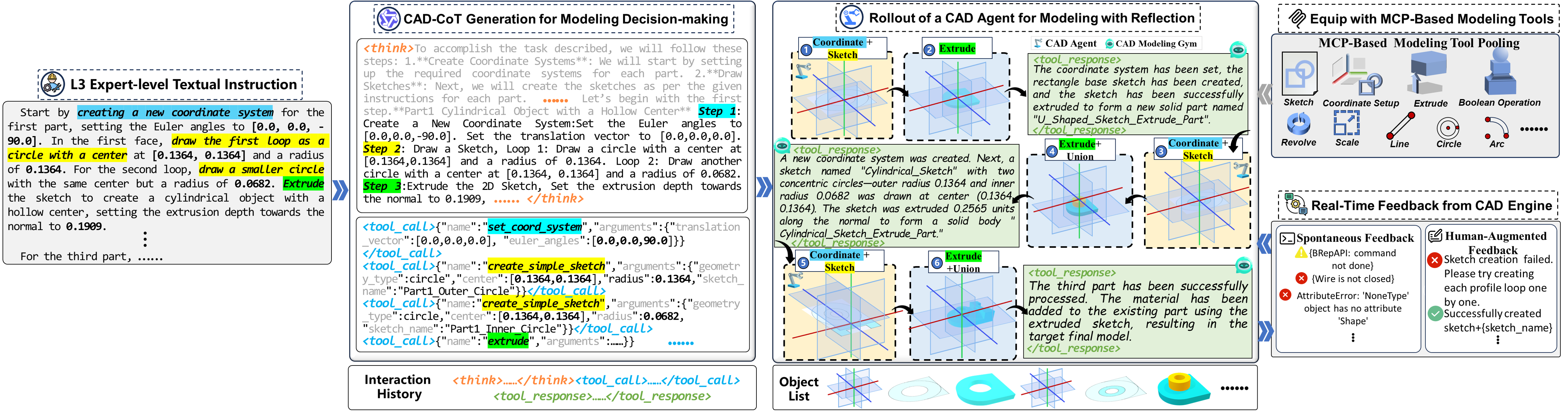}
    \caption{\textbf{CAD-specific Tool-Using Agent Workflow for Text-to-CAD Generation}. Given an expert-level natural language-based CAD design intent, the \textsc{ToolCAD} framework performs 1) modeling decision-making, 2) equip with modeling tools and environment, 3) automatic modeling with reflection. }
    \label{fig:agent}
\end{figure*}

\noindent \textbf{Agentic Reinforcement Learning.} Agent RL has shown promising results in training agents by augmenting LLMs with the ability to invoke external tools for complex tasks \cite{wang2025ragen}, advancing LLMs’ tool-integrated reasoning capabilities in interacting with external environments, such as retrieval engines and code interpret \cite{jin2025search, zhou2025sweet}. Early efforts on agent training explored classical RL algorithms such as DQN \cite{mnih2015human}, and later transitioned to value-based methods like PPO \cite{wang2025acting} and AWR \cite{peng2019advantage} for more stable optimization. Very recent approaches incorporate inference-time search strategy, such as Monte Carlo Tree Search (MCTS) \cite{yuan2025agent}. Besides, training-based approaches employ Direct Preference Optimization (DPO) \cite{rafailov2023direct}, Group Relative Policy Optimization (GRPO) \cite{singh2025agentic, shao2024deepseekmath} to align LLM-based agent rollout trajectory with human preference. 
\section{ToolCAD Workflow}
\subsection{Problem Formulation}
Our goal is to develop an CAD-specific framework for adopting and training tool-using LLM to act as expert-level executors, tailored for various complex text-to-CAD tasks. From the view of language agent task, we model the process of tool-using agents’ autonomous thinking while performing tool-based CAD operations as a finite-horizon Markov Decision Process (MDP) $\mathcal{M = \{S, A, T\}}$, the state $s$ denotes the historical context, which consists of reasoning along with the history of previous modeling actions. 
Given an expert-level instruction $I$, the agent policy $\pi_{\theta}$ must select a tool-based action $a_t$ at any decision step $t$, guided on the current state $s_t$, then decide on the next action to transition to a new state $s_{t+1}$ in a finite-horizon setting. By introducing available real-time feedback from the CAD modeling gym, the tool-using trajectory along with reward path can be easily collected for training. Formally, the reasoning and tool integration of CAD modeling process is defined as:
\begin{equation}
\scalebox{1}{$
a_t\sim\pi_\theta(\cdot|s_t,\tau_{<t},I),\quad(r_t,s_{t+1})\sim \mathcal{P}(\cdot|s_t,a_t)
$}
\end{equation}
where {$\tau_{<t}=\{s_0,a_0,r_0,...,s_{t-1},a_{t-1},r_{t-1}\}$} denotes history interactions including all preceding reasoning-guided tool-calls, observations and rewards. 
\subsection{Framework Overview}
The diagram in Fig.\ref{fig:agent} outlines the workflow of the tool-using LLM agent autonomously executing text-to-CAD tasks within the \textsc{ToolCAD} framework. This framework incorporates three stages: \textbf{1) CAD Modeling Decision-making} – Utilizing chain-of-thought prompting and post-training, enable and evolve the capability of tool-using LLM to plan and decompose complex modeling tasks using CAD modeling Chain of Thought (CAD-CoT) generated in response to L3 expert-level prompts. \textbf{2) Equip with Modeling Tools and Environment} – The agent leverages the CAD-CoT, containing reasoning-guided actions, to call the corresponding custom-designed Model Context Protocol (MCP)-based modeling tools through interactions with the cad environment to advance the process of reasoning and tool integration. \textbf{3) Reflective Automatic CAD Modeling} – At this stage, we employ CAD-specific ReAct to maintain consistency across reasoning-guided actions and the CAD engine. Based on hybrid modeling feedback from the engine, the agent reflects on the outcome of each tool-call, adjusting and executing modeling tools iteratively until the task either succeeds or fails. The detailed implementation of the tool-using agent can be found in the Appendix \ref{apx:A}, \ref{apx:A2}, \ref{apx:A3}. 

\subsection{CAD-CoT Reasoning and Tool Integration}
\textbf{CAD-CoT Prompting.} Expert-level (L3) parametric CAD modeling instructions involve dense numerical parameters (e.g., coordinates, angles, radii, directions…), as well as a standardized and sequential procedural pipeline---spanning Coordinate Setup, Sketching, Extrusion and Boolean Operations (cut, union, common), to create each unit part. Constructing multi-part CAD models challenges tool-using LLMs' long-horizon reasoning and tool integration, which significantly increases the risk of hallucinations in both parameters and action sequences. Hence, in order to elicit the tool-using agents to generate reliable CAD-CoT across multiple parts for complex text-to-CAD generation, we customize the CAD-specific ReAct \cite{yao2023react} prompting strategy to ensure coherence between reasoning and tools. In addition, we instruct the LLM to structure CAD-CoT in a strict reasoning-guided modeling tool-call format, using special tokens (e.g., {\textcolor{thinkblue}{\tt{<think>...}\tt {</think>}}}, {\textcolor{toolgreen}{\tt{<tool\_call>...}\tt {</tool\_call>}}}), to reason out precise and reliable CAD modeling tool chains. 

\subsection{CAD Modeling Gym}
This section presents the RL training components for the CAD tool-using agent, including the real-time hybrid modeling feedback and the trajectory collection pipeline. 

\noindent \textbf{CAD Modeling Feedback Design.} To align text-based outputs with actual CAD execution, we first integrate an interactive and agentic modeling engines with LLMs via the open-source CAD platform FreeCAD by wrapping parametric CAD modeling primitives as agent-callable MCP-based modeling tools. The CAD compiler constitutes the core of the environmental feedback within our gym. 
Following the ReAct interaction paradigm, we design a step-level interactive modeling feedback mechanism from two key perspectives: \textbf{(1) Spontaneous Feedback.} In order to enable the agent receives coarse-grained feedback, we expose the CAD engine's primitive-level tool responses including geometric conflict alerts, constraint warnings, and API error messages. This feedback manifests exceptions or error logs, facilitating the CAD agent to backtrack and self-correct along the modeling trajectory. \textbf{(2) Human-Augmented Feedback.} Because CAD engine's spontaneous feedback mixes code snippets and structured messages, it poorly reflects step-level modeling results, causing potential reasoning and tool-call hallucinations for tool-using agents. 
Therefore, we wrap the modeling tool outputs with human-augmented feedback, producing structured messages labeled as \emph{'success'} or \emph{'fail'}. After each tool invocation, LLM-readable text descriptions of modeling results are stored in the interaction history $s_t$, enabling the agent to perform reflective CAD modeling based on observations $\mathcal{O}_t$. Specifically, the agent uses a geometric object list of actual CAD entities representing the current geometric state of the model to alleviate tool execution hallucinations, rather than relying solely on the recorded interaction history, ensuring reliable reflective modeling.

\begin{figure*}[t!]
    \centering
    \includegraphics[height=5.7cm, keepaspectratio]
    {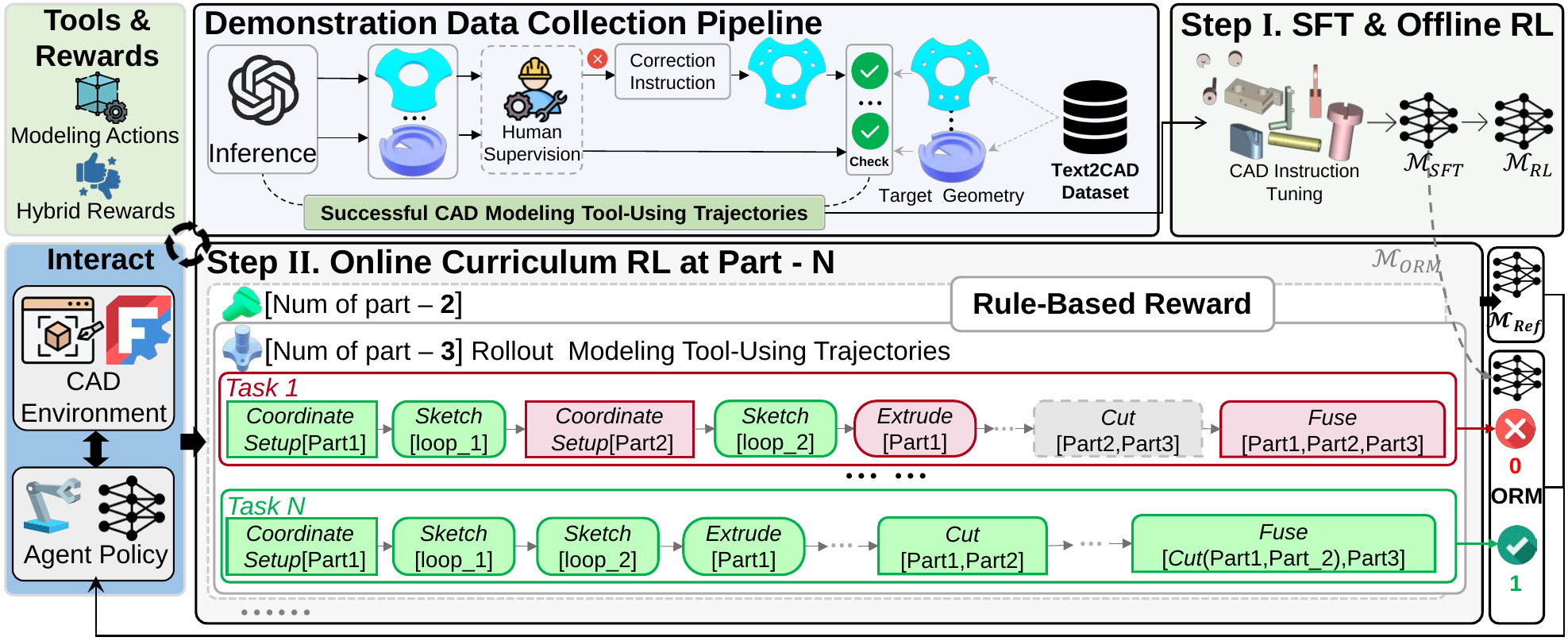}
    \caption{\textbf{Online-RL framework of \textsc{ToolCAD}}. Starting from human-supervised CAD trajectories, the policy is refined into a robust agent through online curriculum reinforcement learning. The ORM training is consistent with human-supervised knowledge.} 
  
    \label{fig:rl}
\end{figure*}

\noindent \textbf{Demonstration Data Collection Pipeline.} 
As illustrated in Fig.\ref{fig:rl}, we construct a demonstration data collection pipeline to produce successful CAD tool-using trajectories from real-time interactions across tasks of varying complexity. Furthermore, to ensure alignment with target geometry, outcome-based evaluation is annotated through human visual check, with correction instructions introduced to guide the agent rollouts toward successful CAD modeling tool-using trajectories. 
This pipeline supports outcome-supervised reward model (ORM) training, enabling automated trajectory-level optimization in RL phases.

\section{Post-training for Evolving CAD Agents} 
\label{sec:training}
\subsection{Reward Modeling.} As mentioned in Section 3.4, the CAD modeling gym provides a hybrid reward modeling mechanism that combines coarse-grained step-level feedback with outcome-supervised signals from the reward model $\mathcal{M}_\mathrm{ORM}$.\\
\textbf{ORM Training.} Even with fine-grained engine feedback, the agent struggles to judge text-to-CAD generation quality and task completion based on the modeling tool-using trajectory. To address this, we fine-tune a LLM as the outcome-supervised reward model $\mathcal{M}_\mathrm{ORM}$ using demonstration data (including negative samples) annotated by human CAD modeling preference, to achieve task success evaluation and provide trajectory-level rewards. Subsequently, we wrap the instruction $I$ and full modeling tool-using trajectory $\tau$ into the prompt to configure $\mathcal{M}_\mathrm{ORM}$ to response "YES" or "NO" to indicate whether a modeling tool-using trajectory successfully completes a target CAD geometry described by instruction $I$, leveraging the learned human knowledge from the language head of $\mathcal{M}_\mathrm{ORM}$. 

At the online reinforcement learning stage, $\mathcal{M}_\mathrm{ORM}$ serves as an automated metrics to access whether the agent's rollout trajectory accomplishes a given task instruction, providing a binary reward signal (0 for failure and 1 for success). The reward is 1 if $\mathcal{M}_\mathrm{ORM}$ assigns a higher probability to “YES” than “NO”, and 0 otherwise.\\
\textbf{Rule-Based Reward.} Beyond encouraging convergence of generated tool-using trajectories to the ground-truth CAD reconstruction supervised by the final outcome reward, we incorporate step-wise rewards by extracting binary signals from feedback after each modeling tool invocation, which contains result labels ("\emph{success}" or "\emph{fail}"), providing coarse-grained reward signals through reward function $\mathcal{R}$:
\begin{equation}
\scalebox{1}{$
\mathcal{R}(s_t, \mathcal{O}_t) = \mathrm{EM}(\mathcal{O}_t,\mathrm{label})
$}
\end{equation}
Moreover, we introduce rule-based format rewards to enforce the tool-using agent policy structure its reasoning, tool use, and CAD environment interactions coherently and reliably, ensuring adherence to the prescribed CAD-CoT prompt template. The format reward function checks the correct order of reasoning and tool-using integration including reasoning ({\textcolor{thinkblue}{\tt<think>}}), tool call ({\textcolor{toolgreen}{\tt<tool\_call>}}), and tool output ({\textcolor{resp-purple}{\tt<tool\_response>}}).
\subsection{Training Strategy.} 
As depicted in Fig.\ref{fig:rl}, we implement a two-stage agentic learning framework tailored for post-training CAD agents, consisting of supervised fine-tuning (SFT) followed by online curriculum reinforcement learning (RL), to evolve strong and robust CAD tool-using agents.\\ 
\textbf{Part-Wise CAD Curriculum Strategy.} Because the unit number in a CAD model critically impacts modeling complexity, a potential challenge in training tool-using LLMs is the instability due to text-to-CAD generation with varying unit counts\cite{du2024blenderllm}. Overlength agent trajectories may lead to context overflow and catastrophic forgetting during rollouts. To stabilize online RL training, we design a part-wise CAD curriculum tool learning strategy that leverages action sequence average perplexity, gradually increasing task complexity by controlling the unit number in each CAD model. For a full trajectory $\tau$, we use the actor $\pi_{\theta}$ to compute its perplexity:
\begin{equation}
\scalebox{1}{$
\mathcal{P}(\tau) = \exp\left( -\frac{1}{T} \sum_{t=1}^{T} \log \pi_\theta(a_t \mid s_t) \right)
$}
\end{equation}
The optimization advances to the next curriculum learning stage once the average perplexity of the held-in test set drops below threshold $\delta$, indicating sufficient policy confidence and proficiency on the current task complexity. Specifically, the perplexity threshold $\delta$ is softly set as a fraction $\alpha$ of the initial threshold, $\delta = \alpha \cdot \delta'$, where coefficient $ \alpha \in(0, 1)$ controls over the exploration-exploitation trade-off in online RL.\\
\textbf{Online RL-Evolving via CAD Exploration.} We first utilize supervised fine-tuning to initialize the base agent policy model, resulting in $\mathcal{M}_\mathrm{SFT}$, using successful CAD modeling tool-using trajectories from static demonstration data.
We adopt online curriculum reinforcement learning across held-out tasks of varying complexity to enable the CAD tool-using agent to self-evolve, addressing imitation learning’s lack of out-of-distribution generalization capability. Specifically, for each rollout modeling task trajectory {$\tau_{i}=\{\tau_{i,(1)},\ldots,\tau_{i,(K)}\}$} of totally {$|\tau_{i}|$} turns and $K$ tokens, trajectory-level optimization enables critic-free training with GRPO strategy to update and its objective is:
\abovedisplayskip=3pt
\belowdisplayskip=2pt
\begin{equation}
\scalebox{0.8}{$
\begin{aligned}
{\mathcal{J}_{\mathrm{GRPO}}(\theta)} 
= \frac{1}{G}\sum_{i=1}^G \frac{1}{|\tau_i|} \sum_{k=1}^{|\tau_i|} 
\min \Bigg[ 
\frac{\pi_\theta(\tau_{i,k}|\tau_{i,<k})}{\pi_{\mathrm{old}}(\tau_{i,k}|\tau_{i,<k})} \cdot \hat{A}_{i,k}, \\
 \mathrm{clip}\left(
\frac{\pi_\theta(\tau_{i,k}|\tau_{i,<k})}{\pi_{\mathrm{old}}(\tau_{i,k}|\tau_{i,<k})}, 
1-\varepsilon, 1+\varepsilon
\right) \cdot \hat{A}_{i,k}-\beta \mathbb{D}_{\mathrm{KL}} \left[ \pi_\theta \| \pi_{\mathrm{ref}} \right]
\Bigg]
\end{aligned}
$}
\end{equation}

where $\varepsilon$ and $\beta$ are hyperparameters, and $\hat{A}_{i,k}$ represent the relative advantage of the $i$-th task trajectory. The clipping threshold $\varepsilon$ ensures stable updates. See more details in Appendix.\ref{apx:B}, and the training process in Appendix.\ref{apx:C3}.

  

\begin{figure*}[ht]
    \centering
    \includegraphics[width=1\linewidth]{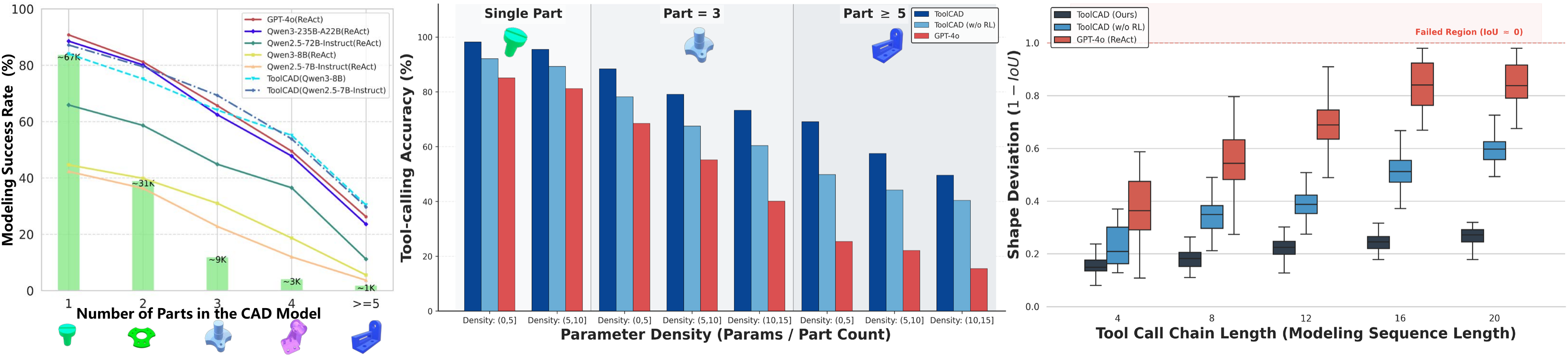}
    \caption{Evaluation of ToolCAD with RL. Modeling success rate, tool-calling accuracy, and geometric precision ($1-\textit{IoU}$) are evaluated to comprehensively measure the tool-using agent's modeling behavior and performance.}
    \label{fig:tools}
\end{figure*}

\section{Experiment}
\label{sec:experiment}
\subsection{Dataset and Evaluation Evironment.} {The effectiveness of prompting and learning framework is evaluated using the \textsc{ToolCAD} environment along with dataset from the DeepCAD and Text2CAD. Text2CAD provides multi-level annotations from $\sim$170K models in the DeepCAD dataset, including four design prompts ranging from abstract to expert levels (L0$\sim$L3). We select and preprocess the L3 expert-level annotations as main text-to-CAD task instructions. In contrast to simple L0-L2, such long and precise L3 instructions are more consistent with industrial-level CAD output standards and agent-friendly. A major limitation of DeepCAD is the scarcity of complex tasks across multiple parts, with most of its $\sim$67K models containing only one part. Therefore, we construct tailored 982 offline demonstration trajectories via GPT-4o as held-in tasks with different levels of complexity for SFT initialization and ORM training. The rest of the dataset is reserved as held-out tasks for online RL training, and 200 test cases from the held-in task serve as the overall evaluation benchmark. More dataset and replication details are in the Appendix.\ref{apx:C1}, \ref{apx:C2}.
\subsection{Baselines} Since there are no existing methods for training specific CAD tool-using agents to perform text-to-CAD modeling, we compare \textsc{ToolCAD} with frontier proprietary LLMs utilizing prompting techniques, as well as open-sourced LLMs trained with alternative methods. For frontier LLMs, we select GPT-4o and Qwen3-235B-A22B, representing the current state-of-the-art in reasoning and agentic tool-use capabilities. For standard open-source models such as Qwen2.5-7B and Qwen3-8B, in addition to leveraging prompt-based reasoning and modeling-tool integration as baselines, we also train them using SFT and off-policy RL---advantage-weighted regression (AWR)\cite{peng2019advantage}, serving as RL-based evolution baselines for \textsc{ToolCAD}. Additionally, Text2CAD, SkexGen\cite{xu2022skexgen} and HNC-CAD\cite{xu2023hierarchical} represent mainstream approaches that generates CAD sequences, providing baselines for evaluating CAD generation quality. We further broadly compare \textsc{TOOlCAD} with prior agentic CAD generation methods (e.g., CAD-Assistant\cite{mallis2024cad}, CAD-Llama\cite{li2025cad}) to highlight its agent-centric design and performance gains. Evaluation metrics are listed in the Appendix.\ref{apx:C4}
\begin{table}[t]
\centering
\resizebox{\linewidth}{!}{
\begin{tabular}{lcccc}
\toprule
\textbf{Model} & \textbf{Method} &  \textbf{IR\(\downarrow\)} & \textbf{MCD\(\downarrow\)}& \textit{Avg.$S_R$(\%)\(\uparrow\)} \\
\midrule
  \rowcolor{gray!30} 
  \multicolumn{5}{c}{\textit{Frontier LLMs}} \\
  \cmidrule(lr){1-5}
\multirow{2}{*}{GPT-4o} 
  &Zero-shot-CoT      & 20.49               & 19.84   & 48.4   \\
  & ReAct              & 9.12    & 5.88 &\underline{62.7} \\
\cmidrule(lr){2-5}
\multirow{2}{*}{Qwen3-235B-A22B} 
  &Zero-shot-CoT      & 30.25              & 37.18 & 51.3 \\
  & ReAct             & 25.65 &29.83 & 60.5 \\
\cmidrule(lr){1-5}
  \rowcolor{gray!30} 
  \multicolumn{5}{c}{\textit{Open-Source LLMs}} \\
  \cmidrule(lr){1-5}
\multirow{2}{*}{Qwen2.5-72B-Instruct} 
  &Zero-shot-CoT       & 52.94             &50.73 &36.1\\
  & ReAct           & 45.17 & 48.29 &43.4 \\
\cmidrule(lr){2-5}
\multirow{5}{*}{Qwen3-8B} 
  & Zero-shot-CoT      & 60.39            &61.78  &18.5         \\
  & ReAct           &54.02  &53.18  &27.9    \\
  & +SFT             &16.13  &29.43  &41.2          \\
  & +AWR             &24.58  &33.64  &37.8            \\
  & +\textsc{ToolCAD}(ours)  &\underline{1.84}   &\underline{1.26}   &61.8  \\
\cmidrule(lr){2-5}
\multirow{5}{*}{Qwen-2.5-7B-Instruct} 
  & Zero-shot-CoT      & 70.32            &68.41   &14.9         \\
  & ReAct           &62.88   &61.39  &23.4        \\
  & +SFT             &15.21   &28.49 &43.7          \\
  & +AWR             &22.74   &30.77 &39.2    \\
  & +\textsc{ToolCAD}(ours)  &\textbf{1.51} &\textbf{ 1.12}  &\textbf{63.9} \\
\cmidrule(lr){1-5}
  \rowcolor{gray!30} 
  \multicolumn{5}{c}{\textit{Transformer-based Generation Models}} \\
  \cmidrule(lr){1-5}
\multirow{1}{*}{DeepCAD} 
  &--      & 15.36              & 13.41       &-- \\
\cmidrule(lr){2-5}
\multirow{1}{*}{TextCAD} 
  &--       & 2.25              & 1.97       &--  \\
\cmidrule(lr){2-5}
\multirow{1}{*}{SkexGen} 
  &--      &4.59              & 3.85 &-- \\
\cmidrule(lr){2-5}
\multirow{1}{*}{HNC-CAD} 
  &--      &3.45              & 3.08 &-- \\
\bottomrule
\end{tabular}
}
\caption{Main comparisons of average CAD modeling success rate(SR) and quality across different baselines. The \textbf{best} and \underline{second-best} models are highlighted.}
\label{main_res}
\end{table}}

\begin{table*}[htbp]
\centering
\resizebox{\linewidth}{!}{
\begin{tabular}{lccccccccc}
\toprule
\textbf{Method} & \textbf{Fine-Tuning} & \textbf{VLM-Based} & \textbf{Backbone} & \textbf{Prompt Type} & \textbf{CAD-Code} & \textbf{ACC}$_T\uparrow$&\textbf{COV}$\uparrow$& \textbf{MMD$\downarrow$} &\textbf{JSD$\downarrow$}\\
\midrule
CAD-Assisant\cite{mallis2024cad} & $\times$ & $\checkmark$(GPT-4o) & GPT-4o & L0 text + docstr + image &$\checkmark$  & 51.25 & 60.32 & 6.13 & 25.09  \\
CAD-GPT\cite{wang2025cad} & $\checkmark$ & $\checkmark$(LLaVA) & LLaVA & L0 text + image &$\times$  &52.78 & 64.59 &4.49 & 22.17 \\
CAD-Llama\cite{li2025cad} & $\checkmark$ & $\checkmark$(CLIP) & LLaMA3-8B-Instruct  & L0$\sim$L1 text & $\checkmark$ &79.46 &74.86 &1.62 & 3.85\\
VideoCAD\cite{man2025videocad} & $\checkmark$ & $\checkmark$(DINOv2/CLIP) & ViT & video frame & $\times$ & --- &49.25 &11.75&32.19\\
FlexCAD\cite{zhang2024flexcad} & $\checkmark$ & $\times$ & LLaMA3-8B-Instruct & L3 text token & $\times$ &\textbf{ 81.43 } &70.49& 2.35 &4.79 \\
CADCodeVerify\cite{alrashedy2024generating} & $\times$ & $\checkmark$(GPT-4/Gemini-1.5-Pro) & GPT-4/Gemini-1.5-Pro/CodeLlama & L2$\sim$L3 text & $\checkmark$ & --- &64.37 & 5.46 & 18.26 \\
Text2CAD\cite{text2cad} & $\checkmark$ & $\times$ & BERT & L0$\sim$L3 text & $\times$ & 60.39 & 61.94 &5.19 &10.28\\
CAD Translator\cite{li2024cad} & $\checkmark$ & $\times$ & Transformer & L0 text & $\times$ &69.28 &65.22 &3.79 &9.14\\
CADFusion\cite{wang2025text} & $\checkmark$(+DPO) & $\checkmark$(GPT-4o) &LLaMA3-8B-Instruct & L1$\sim$L2 text & $\times$ & 72.65 &71.93 &3.07& 5.03\\
CAD-Coder\cite{guan2025cad} & $\checkmark$ (+GRPO) & $\times$ & Qwen2.5-7B-Instruct & L0$\sim$L3 text  & $\checkmark$ & 57.34 &65.48 &5.73 &11.42\\
RLCAD\cite{yin2025rlcad} & $\checkmark$ (+PPO) & $\times$ & Transformer & / & $\times$ & --- & 55.18 &7.45 &8.33\\
\rowcolor{lightpurple}
\textbf{\textsc{ToolCAD}(ours)} & $\checkmark$(+GRPO) & $\times$&  Qwen2.5-7B-Instruct & L3 text & $\times$ & 80.63& \textbf{79.06} &\textbf{1.36} &\textbf{3.21}\\
\bottomrule
\end{tabular}
}
\caption{Comparison of agentic CAD generation methods using vision-language models (VLMs) and large language models (LLMs) across multi-part task with varying complexity.}

\label{tab:agentic_cad_method}
\end{table*}

\begin{figure*}[h]
    \centering
    \includegraphics[width=0.8\linewidth]{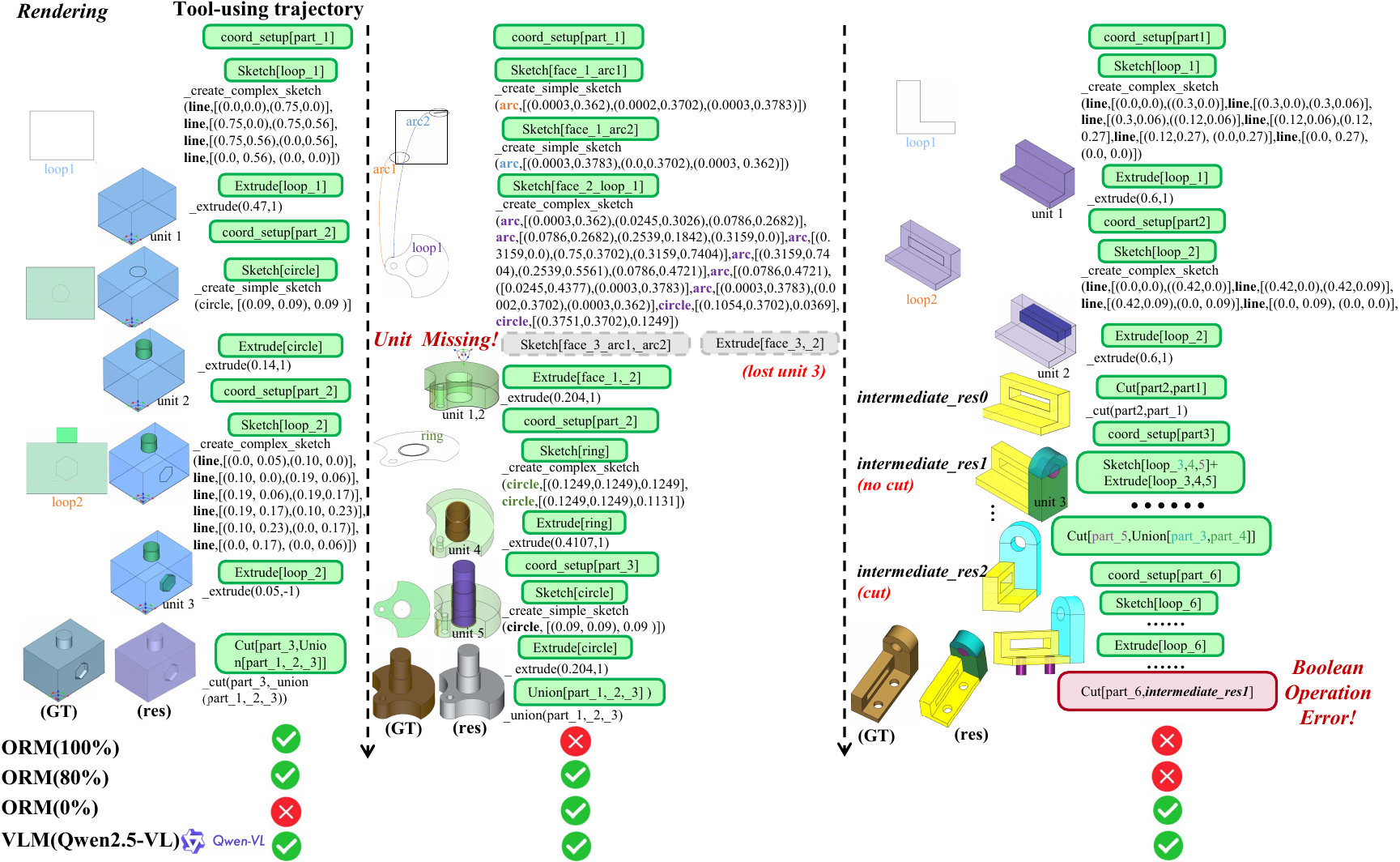}
    \caption{Case study of tool-using agent modeling trajectories. Our ORM (100$\%-$trained) detects subtle step-level near-miss failures such as "\textbf{Unit Missing}" and "\textbf{Boolean Error}" that are visually indistinguishable to the VLM judger (Qwen2.5-VL).}
    \label{vis}
\end{figure*}

\begin{table}[t]
\centering
\resizebox{\columnwidth}{!}{
\begin{tabular}{l l c c c}
\hline
\textbf{Methods} & \textbf{Feedback Source} & \textbf{1P} & \textbf{3P} & \textbf{5+P} \\
\hline
CAD-Assistant (GPT-4o) & Visual+Docstr & 82.5 & 59.1 & 20.8 \\
CAD-GPT (LLaVA) & Visual+Text & 80.3 & 54.8 & 18.4 \\
TOOLCAD-VLM (variant) & Visual+Text & 85.6 & 50.3 & 15.9 \\
TOOLCAD (Ours) & Text & \textbf{87.2} & \textbf{69.3} & \textbf{29.7} \\
\hline
\end{tabular}
}
\caption{VLM-based feedback vs. Engine-based feedback success rate (\%)}
\label{tab:vlm_vs_engine}
\end{table}
\begin{table*}[t]
    \centering
    \small
    \setlength{\tabcolsep}{5pt}
    \begin{tabular}{l c c c c c}
        \toprule
        \textbf{Methods} & \textbf{Instruction-level} & \textbf{COV$\uparrow$} & \textbf{JSD$\downarrow$} & \textbf{Novelty$\uparrow$} & \textbf{IoU$_{best}$$\uparrow$}\\
        \midrule
        \multirow{4}{*}{\textbf{Text2CAD}} 
        & @L0 & 40.27\% & 36.49 & 94.33\% & 40.51\%\\
        & @L1 & 49.35\% & 20.21 & 92.58\% & 45.39\%\\
        & @L2 & 54.68\% & 17.15 & 90.53\% & 54.77\%\\
        & @L3 & 61.94\% & 10.28 & 88.47\% & 61.24\%\\
        \midrule
        \multirow{4}{*}{\textbf{TOOLCAD (Ours)}} 
        & @L0 & 58.19\% & 20.81 & \textbf{98.14\%} & 46.63\%\\
        & @L1 & 61.17\% & 11.25 & 96.31\% & 53.91\%\\
        & @L2 & 70.29\% & 6.47 & 91.75\% & 60.35\%\\
        & @L3 & \textbf{79.06\%} & \textbf{3.21} & 91.62\% & \textbf{65.83}\%\\
        \bottomrule
    \end{tabular}
    \caption{Comparison on geometry quality and diversity across different instruction levels.}
    \label{tab:quality_diversity}
\end{table*}

\subsection{Main Results}
\textbf{Overall Comparisons.} Our main quantitative comparison results, presented in Table~\ref{main_res} and Fig.\ref{fig:tools}, show that Qwen2.5-7B-Instruct and Qwen3-8B trained using \textsc{ToolCAD} achieve average modeling success ratio of 63.9\%, 61.8\% respectively on CAD models with varying unit number, surpassing all SOTA frontier LLM prompting baselines. Meanwhile, GPT-4o (+14.3\%) and Qwen3-235B-A22 (+9.2\%) benefit significantly via \textsc{ToolCAD}'s specific ReAct CAD-CoT prompting, achieving notably higher modeling success rates than with simple prompt-based baseline Zero-shot-CoT. This results highlight that \textsc{ToolCAD} enables LLMs to act as effective tool-using agents for text-to-CAD tasks. For learning-based baselines, the two-stage post-training strategy of \textsc{ToolCAD} outperforms other initial method, such as SFT and offline-RL AWR, remarkably after our offline-to-online RL evolution phase, achieving about 20\% absolute improvement in average success rate on open-source LLMs, from 41.2\% and 43.7\% (SFT) to 61.8\% and 63.9\% (Online RL), demonstrating that online interactive exploration with the CAD engine results in more robust agent policy and mitigates the generalization risk in off-policy learning.\\
\textbf{Performance with Varying Complexity.} As shown in Fig.\ref{fig:tools}, as the part number of CAD model increases, prompt-based tool-using LLMs struggle on multi-part tasks, including GPT-4o. In contrast, online-RL trained models show consistent improvements. For instance (\textit{figure left}), the reasoning-based model Qwen3-8B surpasses GPT-4o on three-part tasks and achieves the best performance among all baselines on 4+ part tasks, attributed to its long-horizon reasoning ability. \textsc{ToolCAD} achieves the highest tool-calling accuracy, particularly in tasks with high parameter density (\textit{figure mid}). This indicates that the \textsc{ToolCAD}'s tool-using agent effectively learns robust instruction-following capabilities in complex modeling, whereas \textsc{ToolCAD} (w/o RL) struggles significantly as parameter complexity scales. To assess progress toward high-quality modeling, we utilize CAD-engine-verifiable metric ($1-\textit{IoU, figure right}$). \textsc{ToolCAD} achieves the minimum  geometric error in long tool call chains, whereas GPT-4o suffers from spatial hallucination as sequences scale. \\
\textbf{Broader Comparison in Agentic CAD Generation.} Table~\ref{tab:agentic_cad_method} shows \textsc{TOOlCAD} outperforms major baselines on multi-part text-to-CAD tasks via L3 text prompt. Most baselines employ VLMs as either judges or feature encoders, but suffer from hallucination and domain mismatch: they appear effective on simple toys yet fail on harder out-of-distribution text-to-CAD tasks. VLM-based methods introduce not transparent feedback that severely undermines robustness. Table~\ref{tab:vlm_vs_engine} demonstrates that VLM-based methods, including our variant ToolCAD-VLM, perform reasonably well on simple geometries but are more prone to accumulating spatial hallucinations during long-horizon reasoning across CAD design tasks of varying complexity (\emph{1P: Simple, 3P: Medium, 5+P: Hard}). \textsc{TOOlCAD} eliminates VLMs entirely, text-driven feedback and agent behavior that is straightforward and stable.
\\
\begin{figure}[ht]
    \centering
     \includegraphics[height=4.2cm, keepaspectratio]{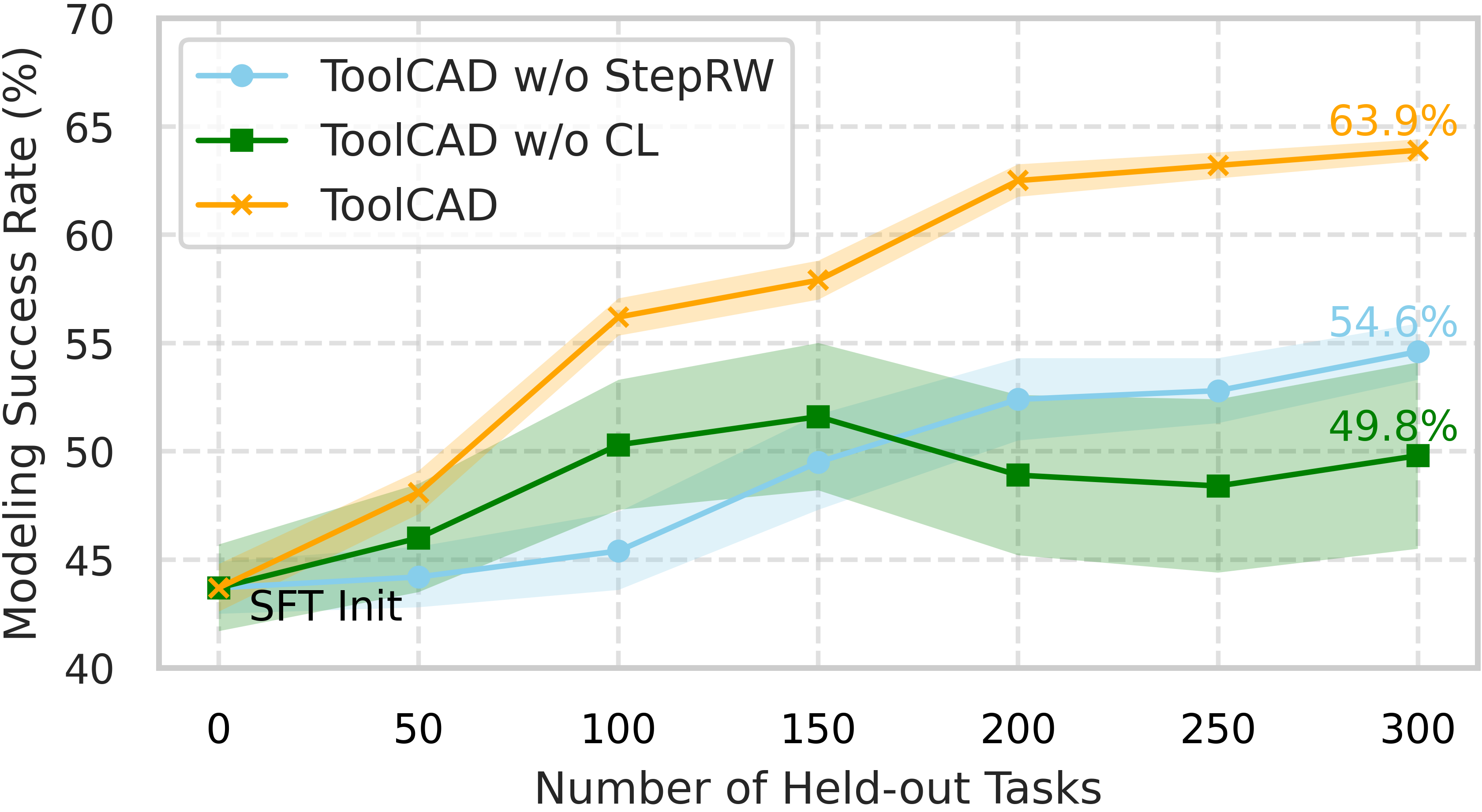}
    \caption{Ablation study of online RL framework on step-level feedback and part-wise curriculum learning strategy.}
    \label{abl}
\end{figure}
\textbf{Instruction-level Geometric Evaluation.} To measure the geometric alignment between the target CAD model and the tool-using agent's output, 
we evaluate the final CAD reconstructions in terms of geometric accuracy and overall quality.
Table~\ref{main_res} reports the quantitative geometric modeling quality of \textsc{ToolCAD} and baseline methods under L3 expert-level instructions. The post-training methods consistently outperform transformer-based generation baselines DeepCAD and Text2CAD. In order to validate \textsc{TOOLCAD}’s intrinsic geometric unfolding capability, we conduct experiments on generative quality and diversity under vague L0/L1 instructions (Table~\ref{tab:quality_diversity}). The results demonstrate that \textsc{TOOLCAD} is not restricted to L3-level precision. Instead, it remains highly competitive under instruction noise by leveraging its learned modeling paradigm to autonomously plan trajectories that achieve the target geometry. Furthermore, we study the effect of CAD agent's the instruction dependence on instruction-level geometric modeling quality using F1 score 
including sketch (\emph{Line}, \emph{Arc}, \emph{Circle}) and extrusion, and the results are summarized in Table~\ref{instruction}. Text2CAD is trained using multi-level instructions from L0$\sim$L3 to realize all skill level consistent modeling. In contrast, our framework generalizes to non-expert instructions via the internalized structure of task trajectories, which empowers the agent to unfold noisy L1 and L2 instructions from abstract intent without explicit tool-call cues.
\subsection{Ablation Studies} 
\begin{table}[ht]
    \centering
    \tiny
    \setlength{\tabcolsep}{3pt}
    \begin{tabular}{l|c|c|c|c|c|c}
        \toprule
          \multirow{2}{*}{\textbf{Method}} & \multicolumn{6}{c}{\textbf{Instruction-level}}\\
             \cmidrule(lr){2-7}
            &\multicolumn{2}{c|}{\textbf{@L1}}&\multicolumn{2}{c|}{\textbf{@L2}}&\multicolumn{2}{c}{\textbf{@L3}} \\
        &Sketch  & Extrusion & Sketch & Extrusion & Sketch & Extrusion\\
         Text2CAD & \textbf{30.35} & \textbf{57.19}& \textbf{47.25} & 69.31 &58.47 &88.62\\
         \textsc{ToolCAD}(Qwen2.5-7B)&24.59  &44.22  &43.23  &75.26 &78.91&93.44\\\textsc{ToolCAD}(Qwen3-8B)& {27.58} & {49.25} & {45.69}& \textbf{77.86}&\textbf{81.52} &\textbf{95.21} \\
         \bottomrule
    \end{tabular}
    \caption{Instruction-level Sketch and Extrusion F1 score.}
    \label{instruction}
\end{table}
\begin{table}[ht] 
    \centering
    \footnotesize 
    \setlength{\tabcolsep}{1pt} 
    \begin{tabular}{lcccc}
        \toprule
        & Our ORM  & GPT-4o & Qwen3-235B & Qwen2.5-VL\\
        \midrule
        Test Cases (\%) & 82.7 & 75.9 & 70.4 & 55.3\\
        Hard Cases (\%) & 65.2 & 54.7 & 50.8 & 41.6\\
        \bottomrule
    \end{tabular}
    \caption{Performance of ORMs.}
    \label{orm}
\end{table}
\textbf{Evaluation of ORM.} In our RL-training framework, ORM (based on Qwen2.5-7B) plays a crucial role in evaluating modeling trajectories to guide the agent’s learning process. Hence, we assess ORM’s effectiveness on the 200 test cases from successful demonstration trajectories and additional failed trajectories, with a particular focus on hard cases involving five or more units ($\geq5$). We compare its performance with several baseline models, including proprietary LLM APIs GPT-4o, Qwen3-235B-A22B and Qwen2.5-VL. Table~\ref{orm} indicates that our ORM  substantially outperforms baselines and retains effective supervision on complex tasks. \\ 
\textbf{Impact of Online RL Optimization Components.} To assess the contribution of key components in our online RL framework, we conduct ablation studies on the step-wise reward mechanism and the part-wise curriculum learning strategy. Accordingly, we implement two variants: \textsc{ToolCAD} w/o StepRW, which removes the step-level rule-based reward function, and \textsc{ToolCAD} w/o CL, which disables the part-wise curriculum learning schedule.

The results, shown in Fig.\ref{abl}, demonstrate that all the components are critical for stability and improvement in the agent’s online exploration. \textbf{(1) The effect of the step-wise reward.} The results indicates that a substantial performance degradation when this rule-based reward function is removed. This highlights the sparse supervision provided by ORM alone is insufficient to support effective optimization of step-level modeling tool-using trajectories. \textbf{(2) The effect of the part-wise curriculum learning strategy.} Compared with \textsc{ToolCAD} w/o CL, 
\textsc{ToolCAD} shows faster progression and consistently higher performance as the number of hold-out tasks increases. Without CL, the agent lacks structured difficulty guidance and struggles when task complexity varies randomly, especially beyond its current skill horizon. Additional analyses are provided in Appendix.\ref{apx:D}\\


    
\section{Case Study}
Fig.\ref{vis} shows three cases in long-horizon multi-part CAD task. Case 1 shows a correct tool call chain for three-part reconstruction, where ORM (100$\%$, 80$\%$) and VLM validator accurately judges success. Cases 2--3 present modeling near-miss failures whose geometry is visually close to ground truth (GT), but the VLM fails to distinguish the subtle visual discrepancies between generated results and GT in the rendered views. Case 2 contains a missing unit failure, detected only by ORM (100$\%$), demonstrating its strength in step-level trajectory supervision. Case 3 illustrates a Boolean Operation error under long tool chaining, highlighting its strong dependence on the tool-using agent’s long-horizon reasoning and planning. The VLM misjudges due to negligible visual cues and subtle blind spots during the CAD design process. More visual results are provided in Appendix.\ref{apx:D4}. 
\section{Conclusion}
In this paper, we propose \textsc{ToolCAD}, a novel LLM-based tool-using agent intelligent CAD system that automates the text-to-CAD modeling framework. This features an interactive CAD gym for agentic reasoning and tool-based interaction, coupled with a curriculum reinforcement learning pipeline that enables LLMs to evolve into proficient CAD tool-using agents. Experimental results suggest that \textsc{ToolCAD} enables open-source LLMs to generalize across complex modeling tasks, supporting their potential as effective backbones for autonomous CAD systems. 
\section*{Limitations} Although empirical experiments have confirmed the effectiveness of the proposed \textsc{ToolCAD}, two major limitations remain. First, while we enable efficient L3 text-prompt-driven tool-using agent workflows for text-to-CAD generation, incorporating visual cue-guided perception as an upstream step to simplify tool calling would be highly valuable. Second, our exploration of CAD-specific tool learning is still limited——lack of self-correction, ORM performance degradation and a limited tool library can hinder task execution. Designing a more reliable Agent–CAD engine interaction framework with stronger geometric feedback may be a more promising future direction than tool learning alone.
\section*{Ethical Statement}We honor the Code of Ethics, and we strictly followed ethical standards in the construction of our dataset. No private data or non-public information is used in our work.

\bibliography{custom}

\clearpage
\setcounter{page}{1}
\appendix
\section{CAD Agent Implementation Details}
\subsection{Prompt Templates}The complete prompt designs for the CAD Tool-Using Agent and the ORM model $\mathcal{M}_\mathrm{ORM}$ are presented below.
\label{apx:A}
\begin{figure*}[ht]
\begin{tcolorbox}[myboxstyle,verbatim,title=\centering CAD Tool-Using Agent Prompt Template]
You are an expert Computer Aided Design (CAD) modeling agent that can fulfill user's high-level instructions. 
Given modeling instructions, you plan the steps of the CAD modeling task and use provided modeling tools to complete the intermediate parts of a final CAD model.  You are provided with modeling tool signatures within <tools></tools> XML tags:\\
\textbf{<tools>}\\
\{modeling tools and descriptions\}\\
\textbf{</tools>}
\\\\
Each part is typically built by:\\
1. Creating a coordinate system.\\
2. Drawing a 2D sketch.\\
3. Extruding it into a 3D shape.\\
4. Optionally applying Boolean Operations.\\
\\
You always first plan the steps of the CAD modeling process by wrapping your reasoning in {\textcolor{thinkblue}{\tt{\textbf{<think>}}}} and {\textcolor{thinkblue}{\tt{\textbf{</think>}}}}.
\\\\
For each function call, return a json object with function name and arguments within {\textcolor{toolgreen}{\textbf{<tool\_call></tool\_call>}}} XML tags:\\
<tool\_call>\\
{"name": <function-name>, "arguments": <args-json-object>}\\
</tool\_call>\\\\
Once all parts are built and assembled into a final model named 'FinalModel', respond with: \textbf{<answer>}COMPLETED\textbf{</answer>}
\\\\
\#REMEMBER:\\
- No need to provide feedback on how to manually complete the modeling operations.\\
- Flexibly use Boolean Operations such as cut out, fuse, and common.\\
\end{tcolorbox}
\end{figure*}

\begin{figure*}[ht]
\begin{tcolorbox}[myboxstyle,verbatim,title=\centering Prompts for $\mathcal{M}_\mathrm{ORM}$ to Assess the Task Trajectory]
You are an expert in evaluating the performance of CAD modeling agent. You will be given a modeling historical context log that memorizes an CAD agent interacting with the CAD environment to automatic CAD modeling, and the designer's intent, the format of history modeling interactions is:\\
\textbf{ \#\#\#}\\
\textbf{Actions}: Action trajectory\\
\textbf{Obverations}: Refletcion\\
\textbf{ \#\#\#}\\\\
The Designer Intent: \{instruction\}\\
Action History: \{modeling histical interaction context\}\\\\
Your goal is to judge whether the agent's execution is successful or not. You must respond with \textbf{YES} or \textbf{NO}.\\\\
\end{tcolorbox}
\end{figure*}
\subsection{Integration of LLM Agents into FreeCAD via the Model Context Protocol}
\label{apx:A2}
\begin{figure*}[ht]
    \centering
    \includegraphics[width=1\linewidth]{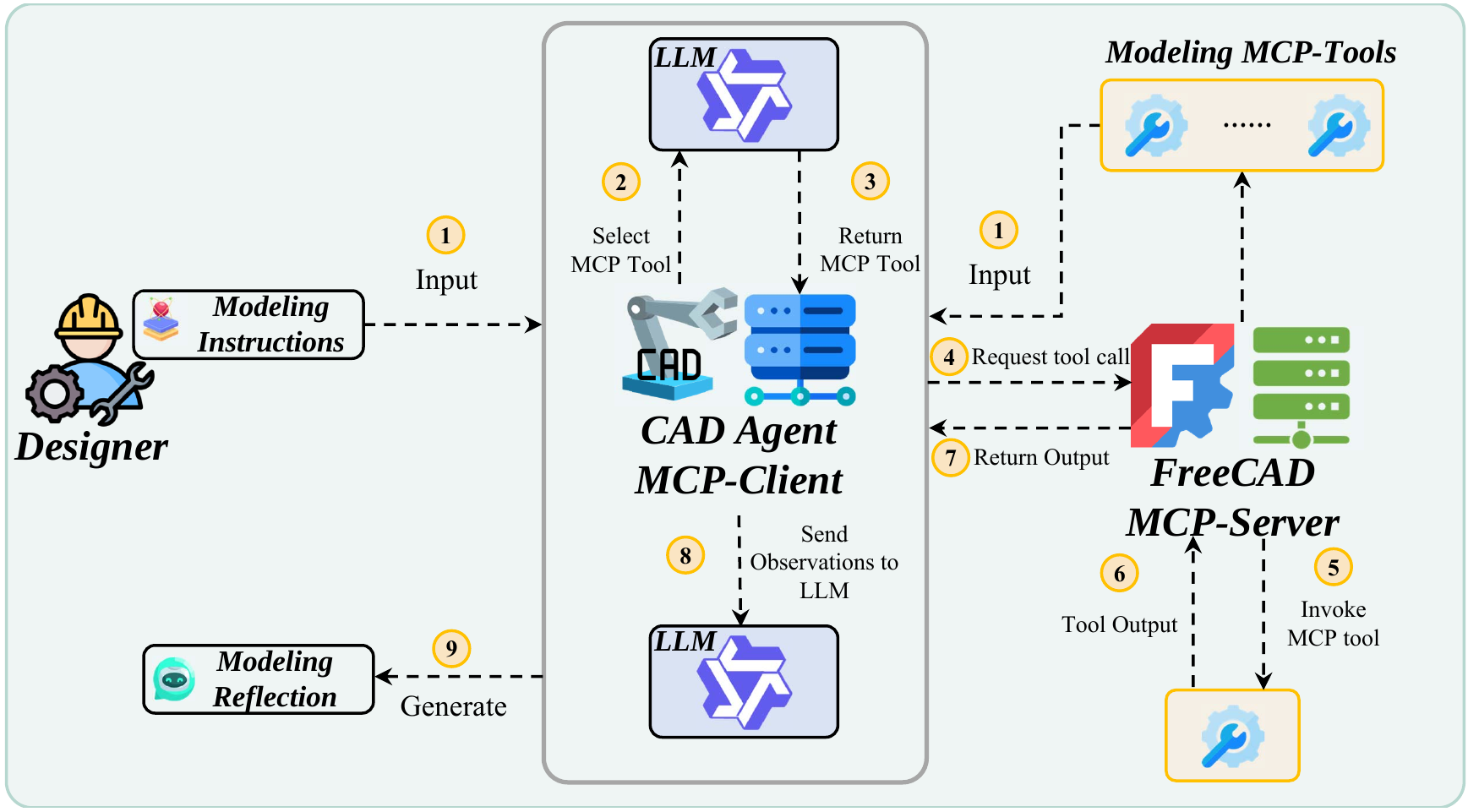}
    \caption{Integration of MCP provides structured environment and tool management, which enhances the extensibility and flexibility of CAD modeling tools.}
    \label{fig:mcp}
\end{figure*}

To efficiently equip the CAD agent with modeling tools and environment, as shown in the \textbf{Figure.\ref{fig:mcp}}, we implement an MCP (Model Context Protocol) Server for the FreeCAD platform to standardize the interfaces and invocation processes of modeling tools. The MCP client facilitates text-to-instruction automatic modeling interactions, enabling the CAD agent to scale and evolve its modeling capabilities.

For custom-designed modeling tools, we implement and encapsulate a set of CAD modeling tools based on the MCP-Server to construct a comprehensive tool library as follows:
\begin{equation*}
\scalebox{0.75}{$
\textsc{Tool Library} = \left\{ 
\begin{aligned}
&\texttt{freecad\text{-}set\_coord\_system}, \\
&\texttt{freecad\text{-}create\_complex\_sketch}, \\
&\texttt{freecad\text{-}create\_simple\_sketch}, \\
&\texttt{freecad\text{-}boolean\_operation}, \\
&\texttt{freecad\text{-}multiple\_fuse}, \\
&\texttt{freecad\text{-}extrude\_face}
\end{aligned}
\right\}
$}
\end{equation*}
\subsection{Feedback Interface Design}
\label{apx:A3}
 The detailed design examples of the Human-Augmented Feedback interface are shown in \textbf{Figure.\ref{fig:create_complex_sketch}} and \textbf{Figure.\ref{fig:bool}}, corresponding to the feedback involved in the primitive-based tool \emph{create\_complex\_sketch} and \emph{bool\_operation}, respectively. 
\section{Details of Policy Update Algorithm in Post-training}
\label{apx:B}
\subsection{Online RL Opimization Details}
We formulate the RL objective function utilizing a CAD engine (CAD environment) $\mathcal{E}_{CAD}$ as follows:
\abovedisplayskip=10pt
\belowdisplayskip=10pt
\begin{equation}
\begin{aligned}
\max_{\pi_{\theta}}\mathbb{E}_{\mathcal{M},I\sim\mathcal{D},\tau\sim\pi_{\theta}(\cdot|I,s;\mathcal{E}_{CAD})}\left[\mathcal{R}_{\phi}(\tau)\right]-\\
\beta\mathbb{D}_{\mathrm{KL}}\left[\pi_{\theta}(\tau\mid I,s;\mathcal{E}_{CAD})\mid\mid\pi_{\mathrm{ref}}(\tau\mid I,s;\mathcal{E}_{CAD})\right]
\end{aligned}
\end{equation}
 $\mathcal{R}_{\phi}$ is the reward function and $\mathbb{D}_{\mathrm{KL}}$ is KL-divergence and  $\beta$ controls regularization. The overall training objective remains the maximization of the expected outcome-based reward $\mathcal{R}_{\phi}(\tau_i)$ from a reference policy $\pi_{ref}$ to maintain stability. For critic-free training leveraging GRPO, the normalized reward $\hat{A}_{i,k}$ is shared across all tokens in $\tau_i$:
\begin{equation}\hat{A}_{i,k}=\frac{\mathcal{R}(\tau_i)-\mathrm{max}(\{\mathcal{R}(\tau_1),\ldots,\mathcal{R}(\tau_G)\})}{\operatorname{std}(\{\mathcal{R}(\tau_1),\ldots,\mathcal{R}(\tau_G)\})}
\end{equation}
where $G$ is the number of modeling trajectories in the batch, hence, $\hat{A}_{i,k}$ denotes the advantage at token $k$ in $\tau_i$.

\subsection{Modeling Behavioral Cloning with Collected Trajectories}
Specifically, we employ behaviroal cloing, also referred to as supervised fine-tuning (SFT), to fine-tune LLM-based CAD tool-using agents by having them mimic the collected modeling trajectories. The training data is derived from  demonstration data collection pipeline, sourced from the our proposed CAD modeling gym. We use these collected trajectories to train a base generally-capable CAD agent with basic instruction-following and CAD modeling abilities. We maxmize the as the initial policy:
\begin{equation}\begin{aligned}
\mathcal{J}_{BC}(\theta) & =\mathbb{E}_{\tau\sim\mathcal{D}}\sum_{t=1}^T\left[\log\pi_\theta(a_t|s_{t-1})\right].
\end{aligned}\end{equation}
where ${\mathcal{D}}$ denotes collected modeling trajectory data.
\subsection{Reward Design}
In TOOLCAD, the final optimization objective for GRPO relies on an aggregated scalar reward $R$, which is a weighted combination of trajectory-level and step-wise signals. Specifically, for each generated trajectory $t_i$ in a group of size $G$, the total reward $R(t_i)$ is defined as:
\begin{equation}
R(t_i) = \alpha \cdot R_{\mathrm{ORM}} 
+ \beta \cdot \left( \frac{1}{T} \sum_{t=1}^{T} R_{\mathrm{step}}^{(t)} \right) 
+ \gamma \cdot R_{\mathrm{format}}
\end{equation}
\begin{itemize}
    \item \textbf{Outcome Reward:} A trajectory-level terminal reward (0 or 1) assessed by the ORM.

    \item \textbf{Step-wise Execution Reward:} For each step $t$, we assign a binary reward $R_{\mathrm{step}}^{(t)} \in \{0, 1\}$. A reward of 1 is granted only if the CAD engine returns a ``Success'' status for the primitive execution. We compute the mean over $T$ steps to maintain a consistent reward scale regardless of trajectory length.

    \item \textbf{Format Reward:} Checks whether the model output contains all required special tokens in the correct order. Specifically, an additional reward of 0.5 is awarded if: (1) all tags are present, (2) the internal order of opening/closing tags is correct, and (3) the overall structure follows:
    \[
    \langle \mathrm{think} \rangle \rightarrow \langle \mathrm{tool\_call} \rangle \rightarrow \langle \mathrm{tool\_response} \rangle.
    \]
\end{itemize}
\section{Demonstration Dataset Details}
\subsection{Collection Pipeline and Data Split}
\label{apx:C1}
In our static demonstration trajectory collection pipeline, different frontier proprietary LLMs (GPT-4o, Qwen3-235B-A22B, Gemini-1.5-pro, Qwen3-32B) are employed to perform agent inference for various text-to-CAD tasks, aiming to generate a diverse distribution of modeling tool-using trajectories. Each generated trajectory is visually checked and aligned with the corresponding ground-truth CAD model under expert supervision. When current geometry errors or omissions are identified, correction instructions are provided to prompt the tool-using agent in correcting CAD-CoT and completing missing steps, thereby ensuring the overall correctness of the trajectories. In total, 982 successful modeling-using trajectories with varying complexity are collected.  
\begin{table}[ht]
\centering
\begin{tabular}{lccr} 
\toprule
\textbf{Number of Part} & \textbf{\#Train Traj.} & \textbf{\#Test Traj.} \\
\midrule
Part-1                  & 100           & 40     \\
Part-2                  & 130           & 40     \\
Part-3                  & 300           & 40        \\
Part-4                  & 130           & 40        \\
Part-$\ge$5             & 122           & 40        \\
\midrule
Total                   & 782           & 200     \\
\bottomrule
\end{tabular}
\caption{The statistics of held-in modeling tool-using trajectories.}
\label{tab:sat}
\end{table}
\begin{figure}[t]
    \centering
    \includegraphics[width=\linewidth]{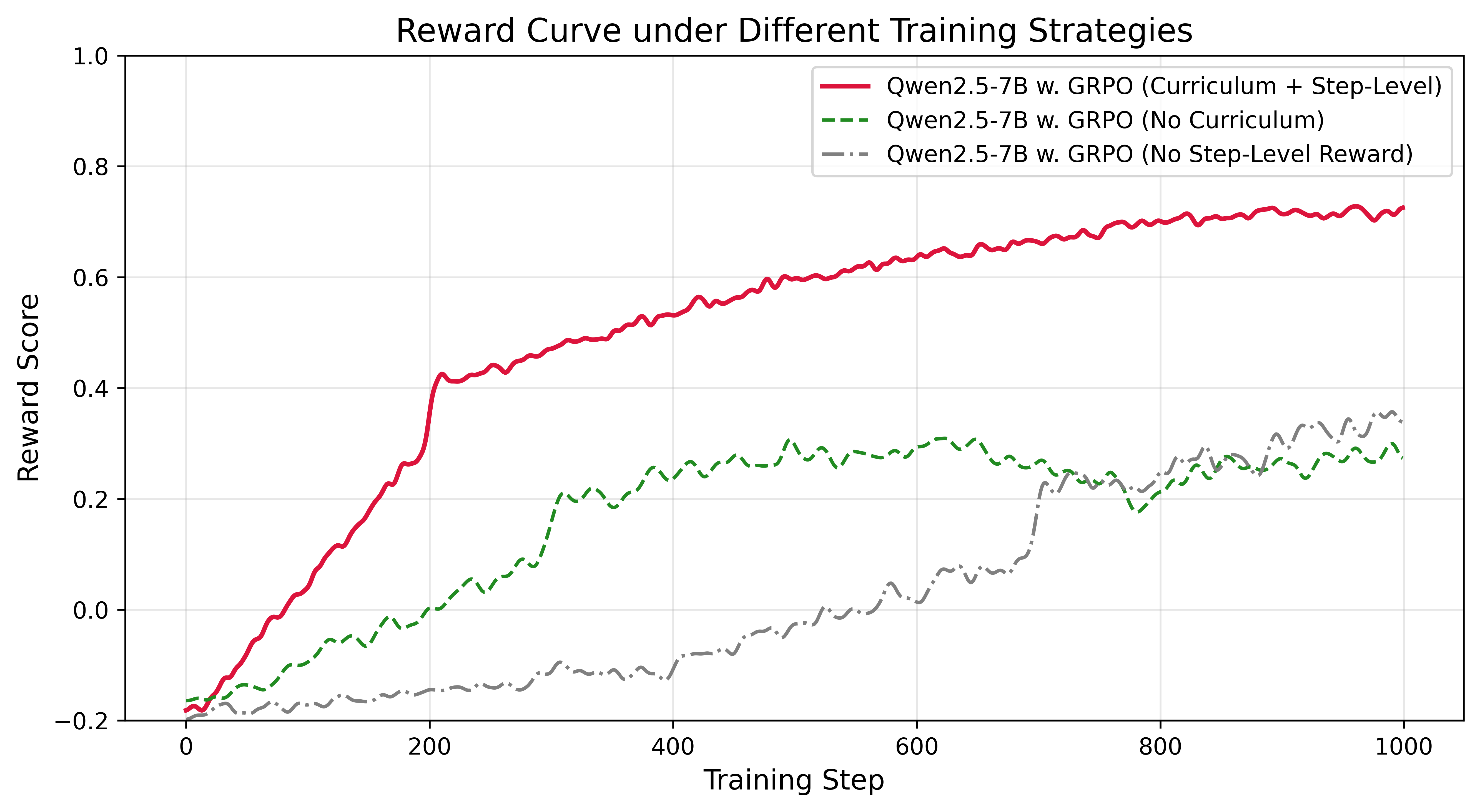}
    \caption{Comparison of different training strategies.}
    \label{fig:reward}
\end{figure}
To ensure a balanced evaluation across tasks of varying complexity, we perform stratified sampling when partitioning the 982 collected trajectories into training and test sets. Specifically, we divide the data into five complexity levels (Part-1 through Part-$\ge$5), and allocate a fixed number of 40 trajectories per level to the test set. The remaining trajectories are assigned to training, resulting in 782 training trajectories and 200 test trajectories. This stratification guarantees that the test set remains uniformly distributed across complexity tiers, while preserving sufficient samples per tier in the training set for effective learning. The detailed statistics are shown in Table~\ref{tab:sat}. A subset of 200 trajectories is selected from the 982 collected demonstrations as test cases for held-in tasks, serving as evaluation benchmarks for SFT, online RL, and ORM models.
\subsection{Implementation Details}
\label{apx:C2}
All experiments are conducted using 8 $\times$ H20-96GB GPUs. The detailed training process can be found in Algorithm \ref{alg:online_curriculum_rl}. The learning rate for iterative SFT is $1e-5$, with 0.1 warm-up ratio and a cosine scheduler. In online RL phase, we sample 8 rollouts per modeling task instruction with a temperature of 0.1. We set a high generation budget of 8,192 tokens to accommodate long-horizon, multi-step reasoning and tool-using. The maximum context window is set to 16384 tokens, and the maximum response length per rollout is 512 tokens. Training is performed using GRPO with a batch size of 32 and a learning rate of $2e-6$. Our codebase is built on Verl and LLaMA-Factory. 
\subsection{Training Procedure}
\label{apx:C3}
The overall pipeline of the proposed algorithm is illustrated in Algorithm.\ref{alg:online_curriculum_rl}.
\subsection{Metrics}
\label{apx:C4}
To thoroughly evaluate \textsc{ToolCAD}'s generation quaility and the modeling performance of the tool-using agent, we use these metrics:
\begin{itemize}
    \item \emph{Invalidity Ratio} (IR): quantifies the percentage of the output CAD models that fail to be converted to point clouds.
    \item \emph{Chamfer Distance} (CD): calculates point-wise proximity between point clouds sampled from $\mathcal{X}$ and $\mathcal{Y}$:
\begin{equation}
\begin{split}
\mathrm{d}_{CD}(\mathcal{X},\mathcal{Y}) &= \frac{1}{|\mathcal{X}|} \sum_{x \in \mathcal{X}} \min_{y \in \mathcal{Y}} \|x - y\|_2^2 \\
&\quad + \frac{1}{|\mathcal{Y}|} \sum_{y \in \mathcal{Y}} \min_{x \in \mathcal{X}} \|y - x\|_2^2
\end{split}
\end{equation}
    \item \emph{Minimum Matching Distance} (MMD): quantifies the average distance between the generated model and its closest-matching reference shape.
    \item \emph{Intersection over Union} (    IoU): as the foundational metric for measuring global volumetric alignment between the generated model $\mathcal{G}$ and the reference model $\mathcal{S}$.
    \begin{equation}
\text{IoU}(\mathcal{G}, \mathcal{S}) = \frac{\mathcal{G} \cap \mathcal{S}}{\mathcal{G} \cup \mathcal{S}},
\end{equation}
    \item \emph{Parameter Density}: This metric calculates the average complexity per shape, defined as:
\begin{equation}
PD = \frac{N_{param}}{N_{unit}},
\end{equation}
$N_{param}$ denotes the total parameter count (e.g., coordinates, radii, and angles) and $N_{unit}$ represents the total number of parts in a CAD model.
\item \emph{Coverage} (COV): measures how well the generative model covers the real data distribution.
{\small{
\begin{equation}
\text{COV}(\mathcal{G}, \mathcal{S}) = \frac{|\{\arg \min_{\mathcal{Y} \in \mathcal{S}} d_{CD}(\mathcal{X}, \mathcal{Y}) | \mathcal{X} \in \mathcal{G}\}|}{|\mathcal{S}|}
\end{equation}}}
\item \emph{Jensen-Shannon Divergence} (JSD):
measures the dissimilarity between two point clouds from the perspective of voxel distribution.
{\small{
\begin{equation}
\text{JSD}(P_{\mathcal{G}}, P_{\mathcal{S}}) = \frac{1}{2} D(P_{\mathcal{S}} \parallel M) + \frac{1}{2} D(P_{\mathcal{G}} \parallel M),
\end{equation}}}
where $M = \frac{1}{2}(P_{\mathcal{S}} + P_{\mathcal{G}})$ and $D$ is the KL-divergence. $P_{\mathcal{G}}$ and $P_{\mathcal{S}}$ are distributions of points in the generated and reference models, respectively.
\item \emph{IOU\textsubscript{best}}: We adopt the $\mathrm{IoU}_{\text{best}}$ metric proposed in~\cite{doris2026cad} for evaluating solid geometry similarity. 
\end{itemize}

\section{More Quantitative Result}
\label{apx:D}
\subsection{Training Process} Figure.\ref{fig:reward} compares three reinforcement-learning strategies.
Curriculum + Step-Level Reward (red) achieves the fastest and most stable improvement, showing the benefit of combining dense step-level feedback with progressive task difficulty.
No Curriculum (green), which retains ORM and step-level rewards but removes difficulty scheduling, converges more slowly and exhibits higher variance.
In contrast, No Step-Level Reward (gray), which relies solely on a final binary ORM reward, performs poorly due to sparse feedback and unstable credit assignment.
These results highlight that step-level supervision is critical for stable learning, while curriculum further enhances convergence.
\begin{figure}[!t]
    \centering
    \includegraphics[width=1\linewidth]{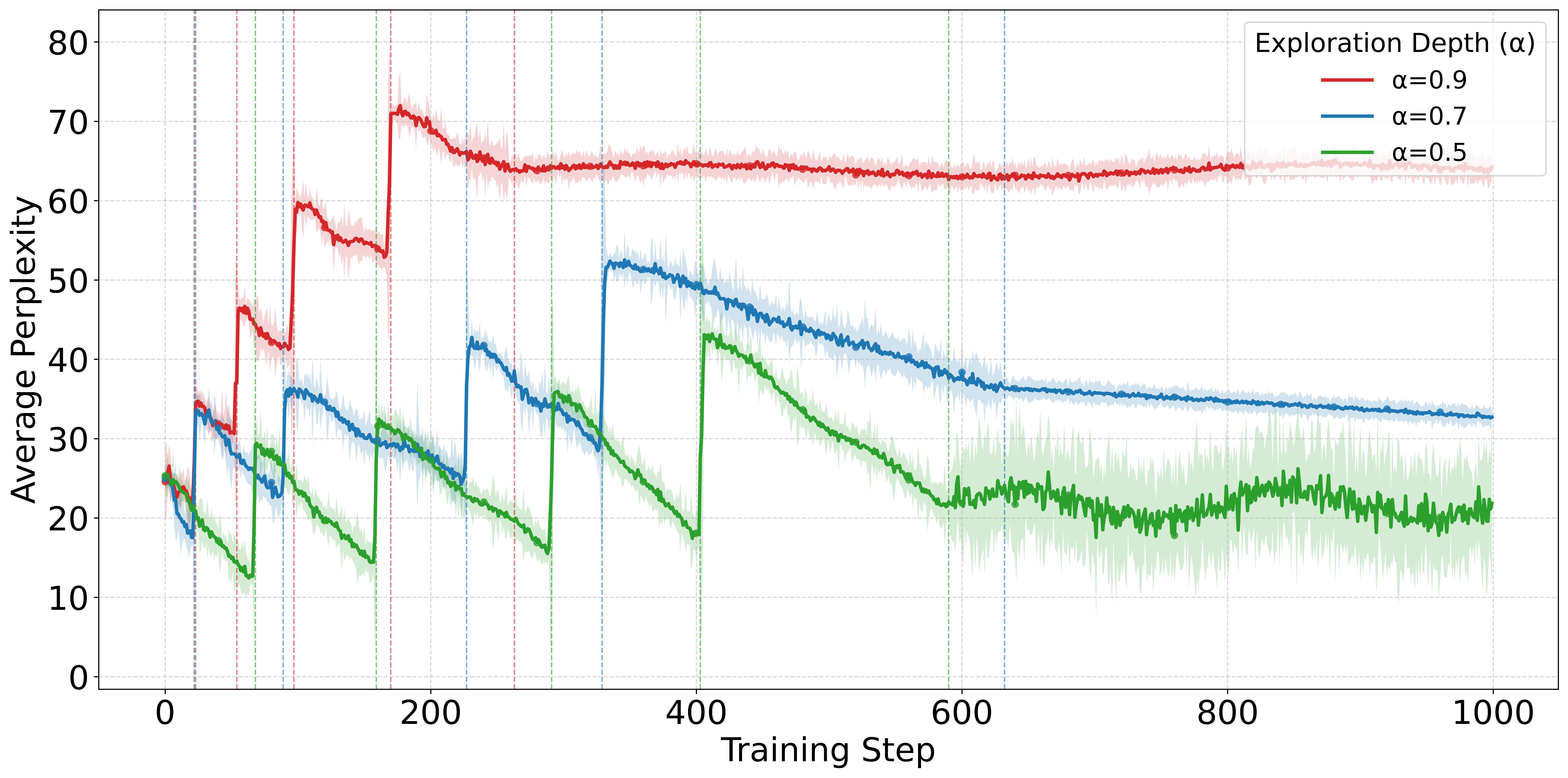}
    \caption{ Average perplexity vs. threshold factor $\alpha$ during online exploration.}
    \label{ppl}
\end{figure}
\subsection{Extra Ablations} 
\textbf{The impact of perplexity.} We investigate how the coefficient $\alpha$ influences average perplexity during online exploration on held-out modeling tasks, providing insights into its role in modulating exploration depth and the underlying exploration-exploitation trade-off for training proficient CAD agents, as shown in Fig. \ref{ppl}. The study is conducted on curriculum learning strategy with three threshold coefficients $\alpha = 0.9, 0.7, 0.5$. When $\alpha = 0.9$, the agent quickly reaches final-stage tasks (part count $\geq$ 5) due to low exploration depth, but maintains high perplexity and fails to improve, indicating insufficient skill acquisition. In contrast, with $\alpha = 0.5$, he agent explores more deeply but with higher time consumption. It shows sharp perplexity spikes on hard tasks, indicating instability of learning likely due to task difficulty gaps and ORM degradation. When $\alpha = 0.7$, the agent maintains moderate and slightly decreasing perplexity, suggesting a balanced trade-off and better alignment with the test distribution, leading to more stable training.
\begin{table}[!t]
\centering
\small
  \setlength{\tabcolsep}{3pt}

\begin{tabular}{lcc}
\toprule
\textbf{Model} & \textbf{Avg Tokens} & \textbf{Avg Latency (ms)} \\
\midrule
\textsc{ToolCAD}(Qwen2.5-7B) & 5178 & 648 \\
\textsc{ToolCAD}(Qwen3-7B) & 6493 & 788 \\
GPT-4o & 11703 & 982 \\
Qwen3-235B-A22B & 9820  & 1108 \\
CAD-Assisant(GPT-4o) & 30802 & 1326 \\
\bottomrule
\end{tabular}
\caption{Comparison of average token usage per tool-call and average tool-call latency across different models.}
\label{tab:toolcall_comparison}
\end{table}


\subsection{Complexity}
Table~\ref{tab:toolcall_comparison} compares average token usage per tool call and tool-call latency across models. Reasoning-oriented models (e.g., GPT-4o, Qwen3-235B-A22B) consume substantially more tokens and incur higher latency, reflecting their stronger planning and multi-step reasoning behaviors. In contrast, \textsc{ToolCAD}(Qwen2.5-7B-Instruct) achieves the lowest latency (648 ms) with moderate token usage, showing better efficiency but limited reasoning depth. The increasing token–latency trend suggests a trade-off between reasoning capacity and interaction efficiency in tool-using CAD workflows, where larger reasoning models excel at complex modeling but introduce non-trivial computational overhead.
\subsection{Visual Qualitative Results}
\label{apx:D4}
\begin{figure*}[ht]
  \centering
\caption{ToolCAD's Agent Modeling Examples}
  \label{fig:my_label}
  \includegraphics[width=1\linewidth]{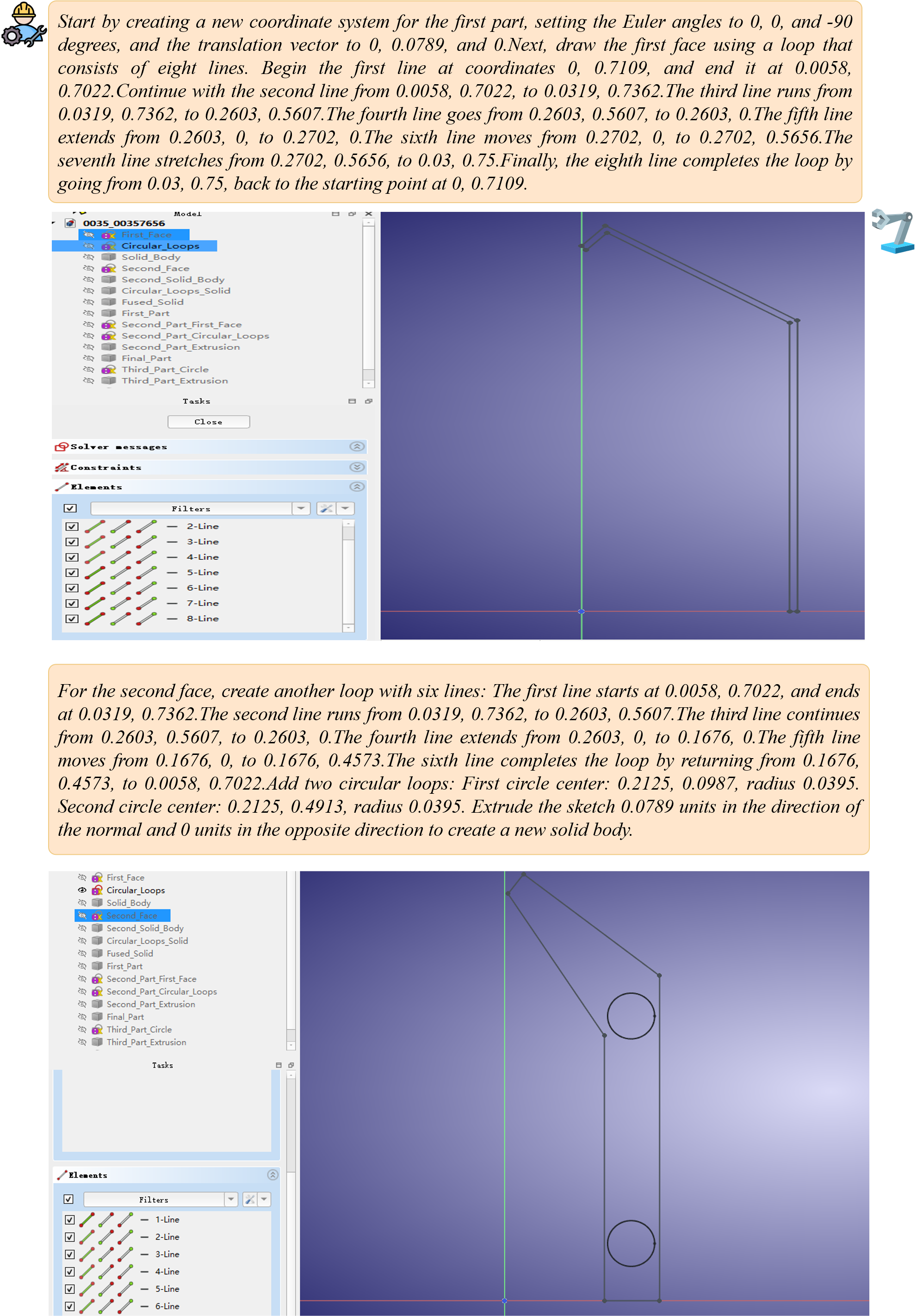}
\end{figure*}

This section provides complete tool-using agent modeling trajectories for text-to-CAD generation across tasks with varying part counts. The cases include the full tool-calling sequence with model-generated part naming, Boolean Operations (Union/Cut), and detailed CAD construction logs, illustrating intermediate geometry and key modeling details. \textsc{ToolCAD}’s agent modeling examples are presented in Figure.\ref{fig:my_label}.
\begin{figure}[ht]
    \centering
    \includegraphics[width=1\linewidth]{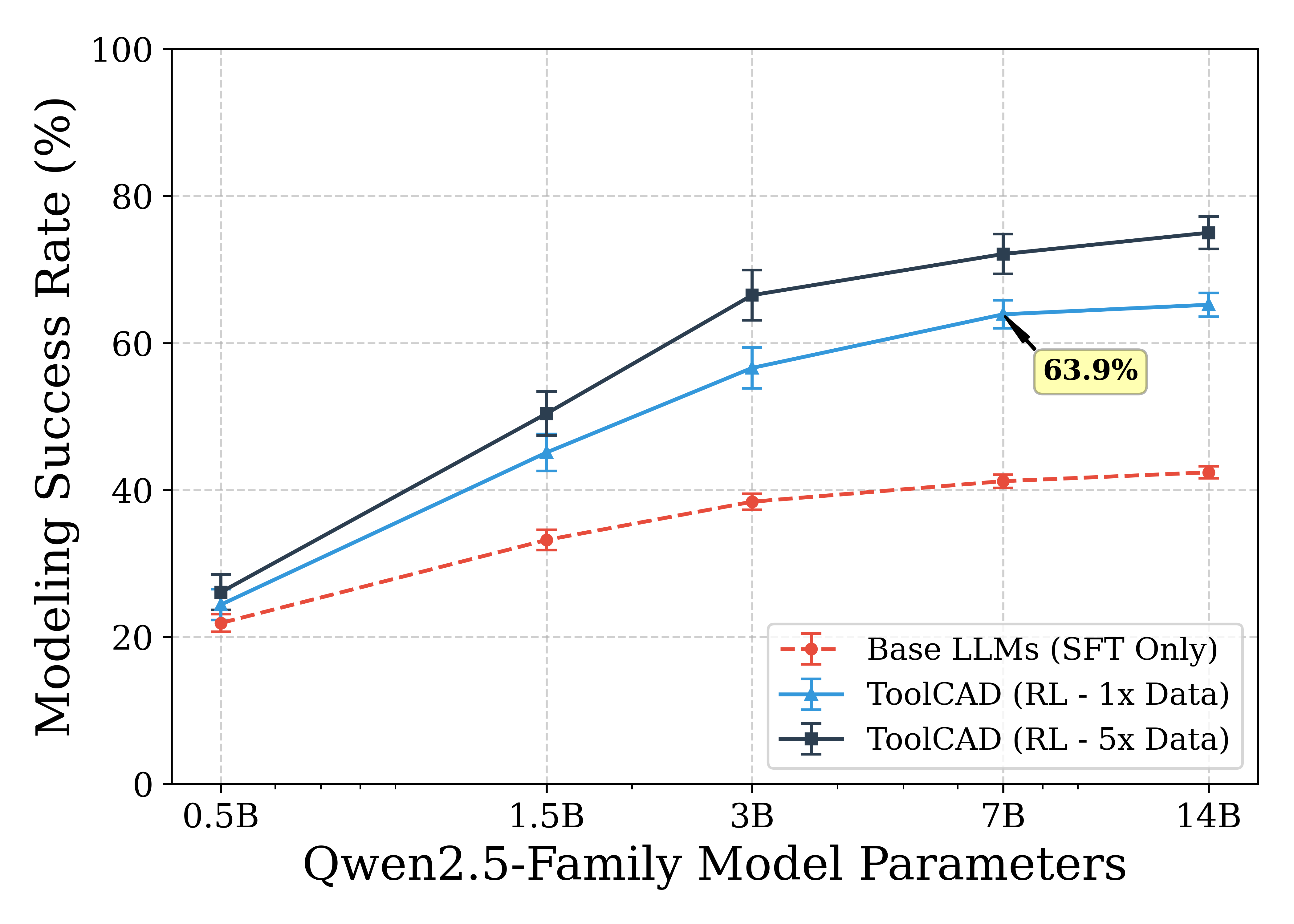}
    \caption{Scaling Laws of CAD Tool-using Agents.}
    \label{scale}
\end{figure}
\subsection{Other Analysis}
\label{apx:D5}
To investigate the performance boundaries of CAD modeling agents, we evaluate scaling behaviors across the Qwen2.5 model family, ranging from 0.5B to 14B parameters. Our analysis focuses on the trade-off between model parameter scale and online reinforcement learning (RL) exploration scale.\\ 
\textbf{The SFT Performance Plateau.} 
As illustrated in the red dashed curve (see Figure.\ref{scale}), base models via SFT exhibit a rapid saturation effect. While increasing parameters from 0.5B to 3B yields noticeable gains, the performance enters a plateau beyond the 7B scale, with the success rate stagnating. This suggests that traditional supervised scaling is insufficient for mastering the long-horizon tool-call chains required for professional-grade CAD modeling.\\
\textbf{RL-Scale as the Primary Performance Driver.} Across all model sizes, applying RL consistently outperforms the SFT-only baseline, confirming the importance of RL for structured reasoning and tool-augmented CAD modeling. More critically, increasing online RL training data from 1× to 5× leads to substantially larger gains than simply increasing model parameters. The 5× RL curve shows a clear upward shift across scales, demonstrating that data scaling, rather than model scaling, is the primary bottleneck under the current RL regime. For CAD tool-using agents, online RL data scaling substantially improves modeling success, but the gains are not unbounded—RL training instability throttles the effective scaling range and ultimately bounds the performance ceiling.

\begin{figure*}[ht]
  \centering
  \includegraphics[width=1\linewidth]{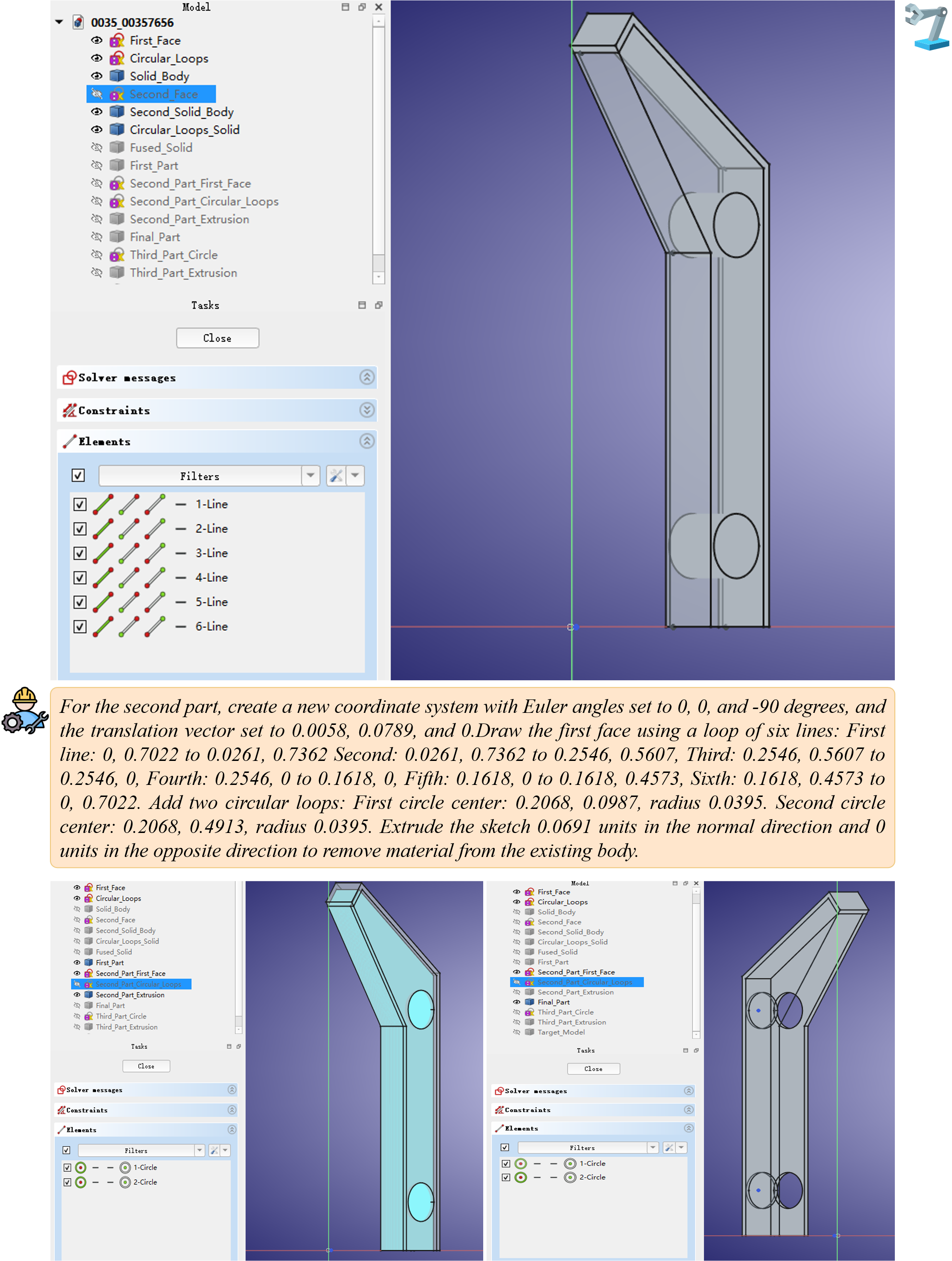}
\end{figure*}

\begin{figure*}[ht]
  \centering
  \includegraphics[width=1\linewidth]{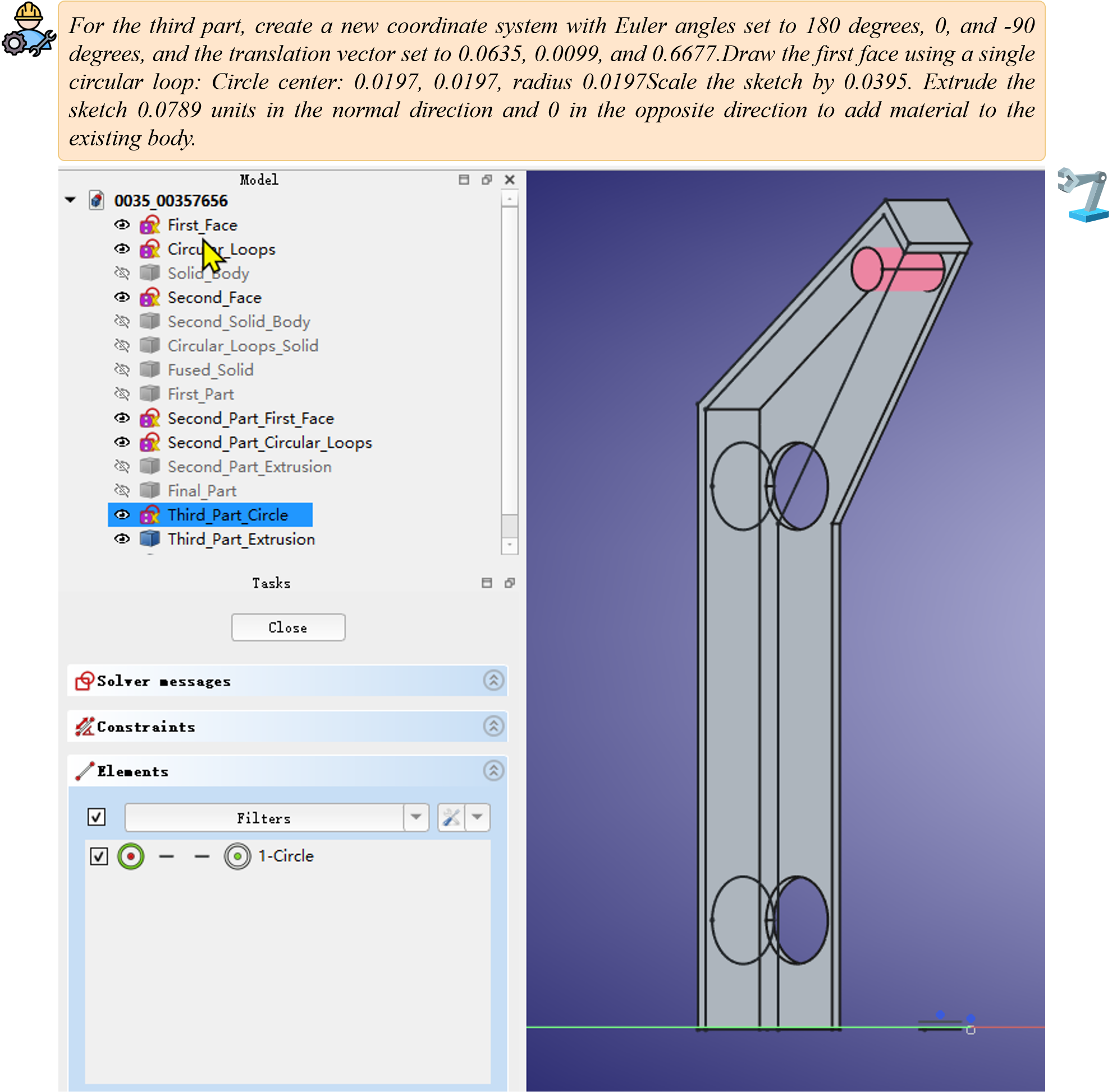}
\end{figure*}

\begin{figure*}[ht]
  \centering
  \includegraphics[width=1\linewidth]{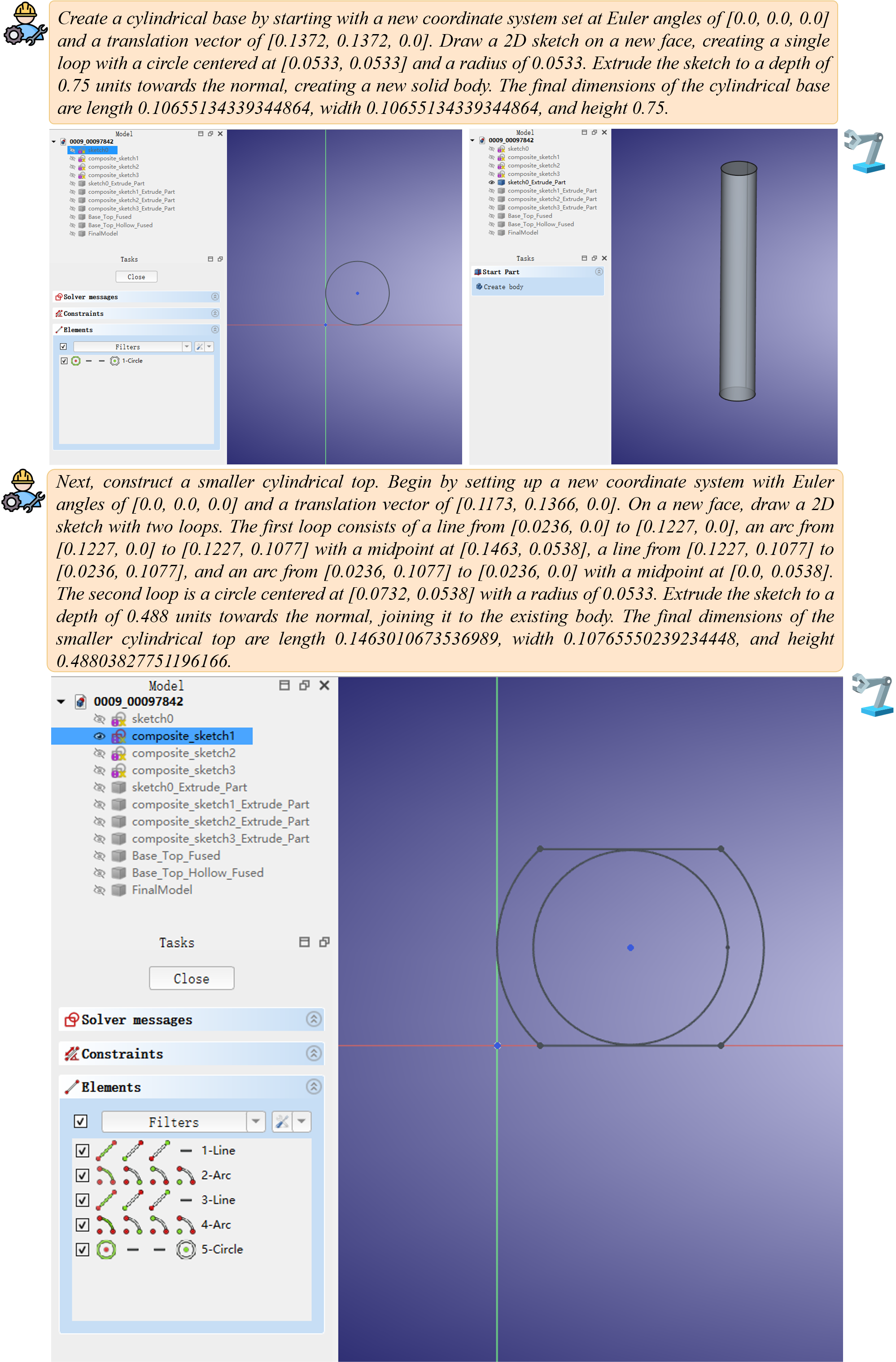}
\end{figure*}

\begin{figure*}[ht]
  \centering
  \includegraphics[width=1\linewidth]{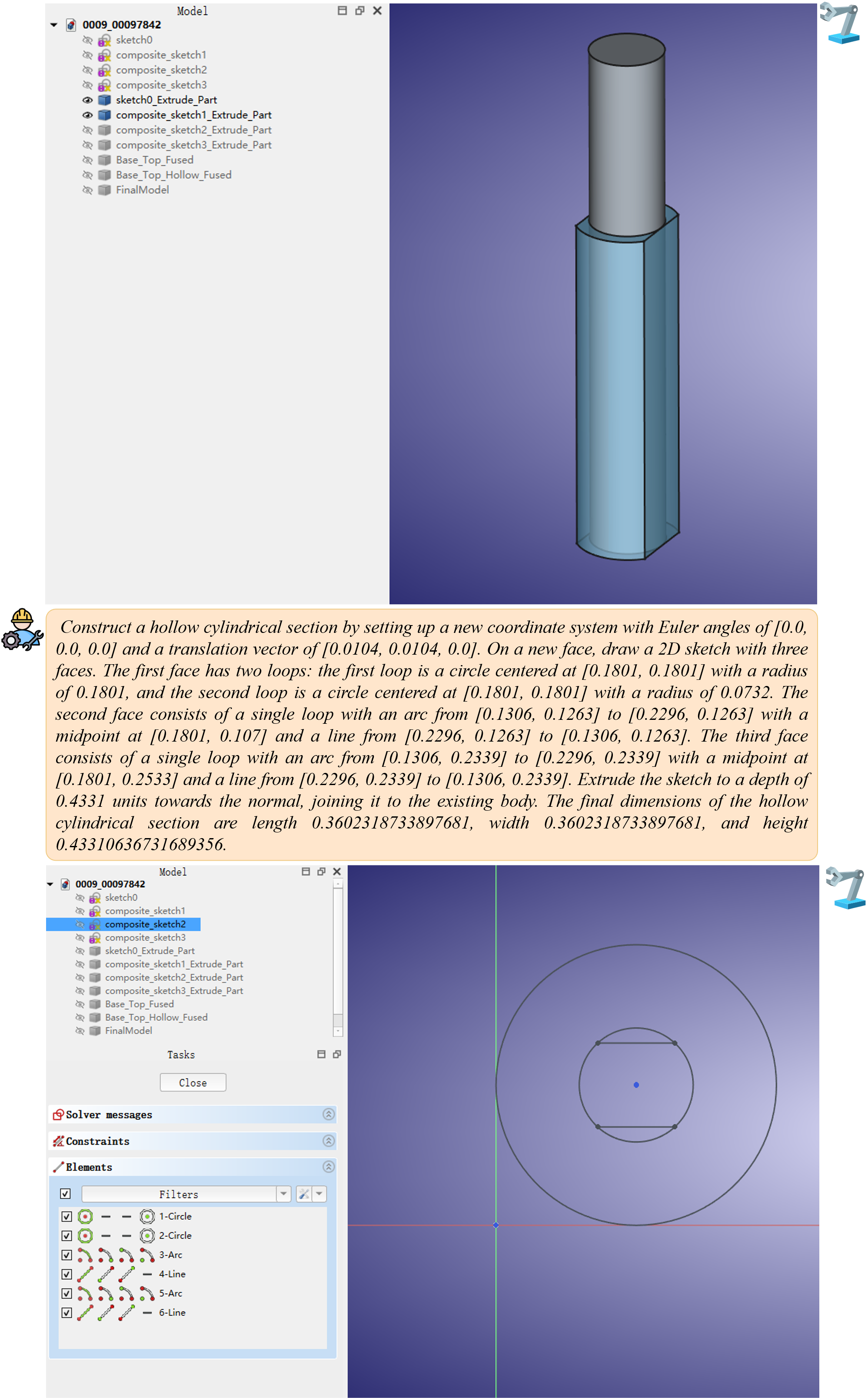}
\end{figure*}

\begin{figure*}[ht]
  \centering
  \includegraphics[width=1\linewidth]{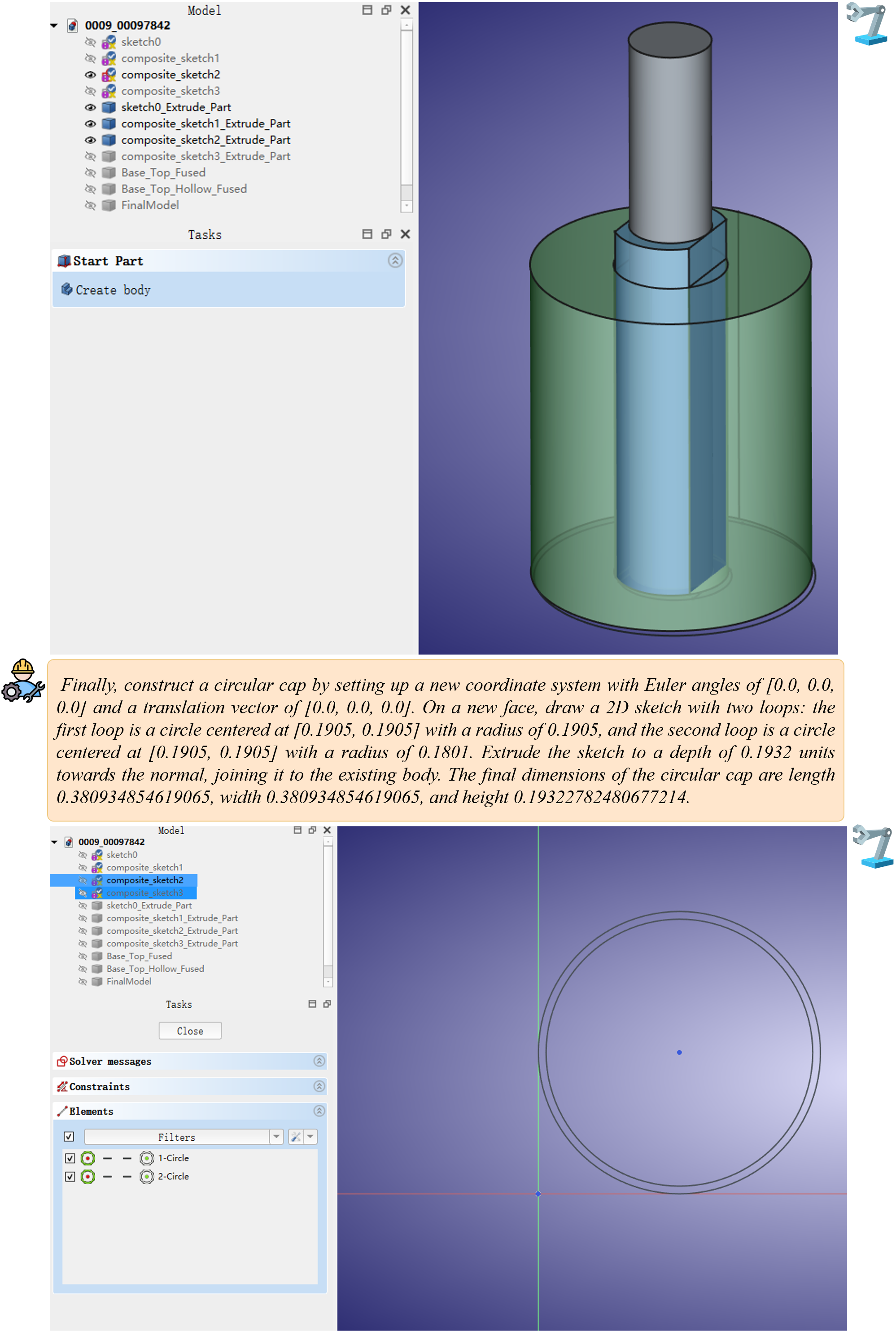}
\end{figure*}

\begin{figure*}[ht]
  \centering
  \includegraphics[width=1\linewidth]{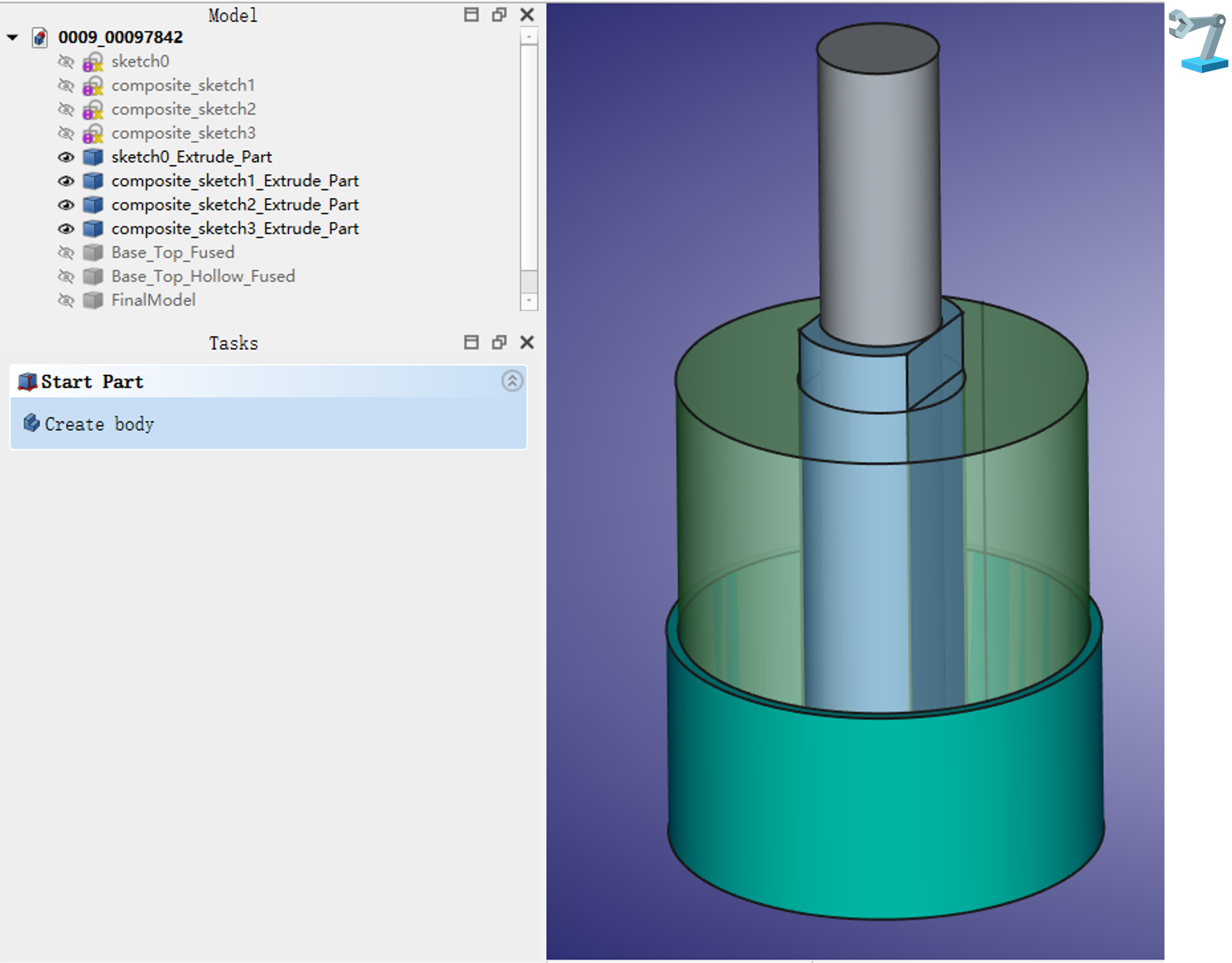}
\end{figure*}

\begin{figure*}[ht]
  \centering
  \includegraphics[width=1\linewidth]{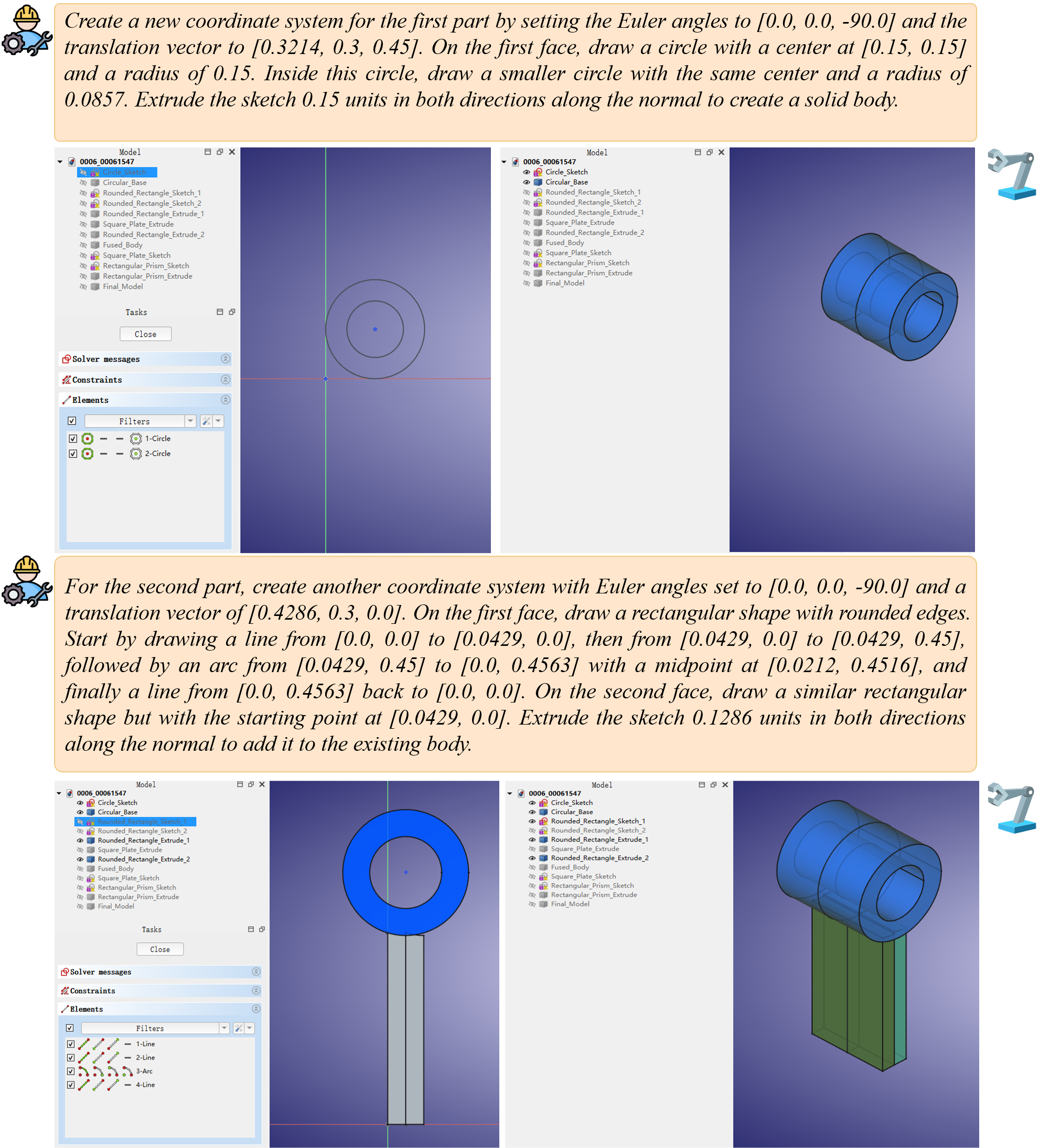}
\end{figure*}

\begin{figure*}[ht]
  \centering
  \includegraphics[width=1\linewidth]{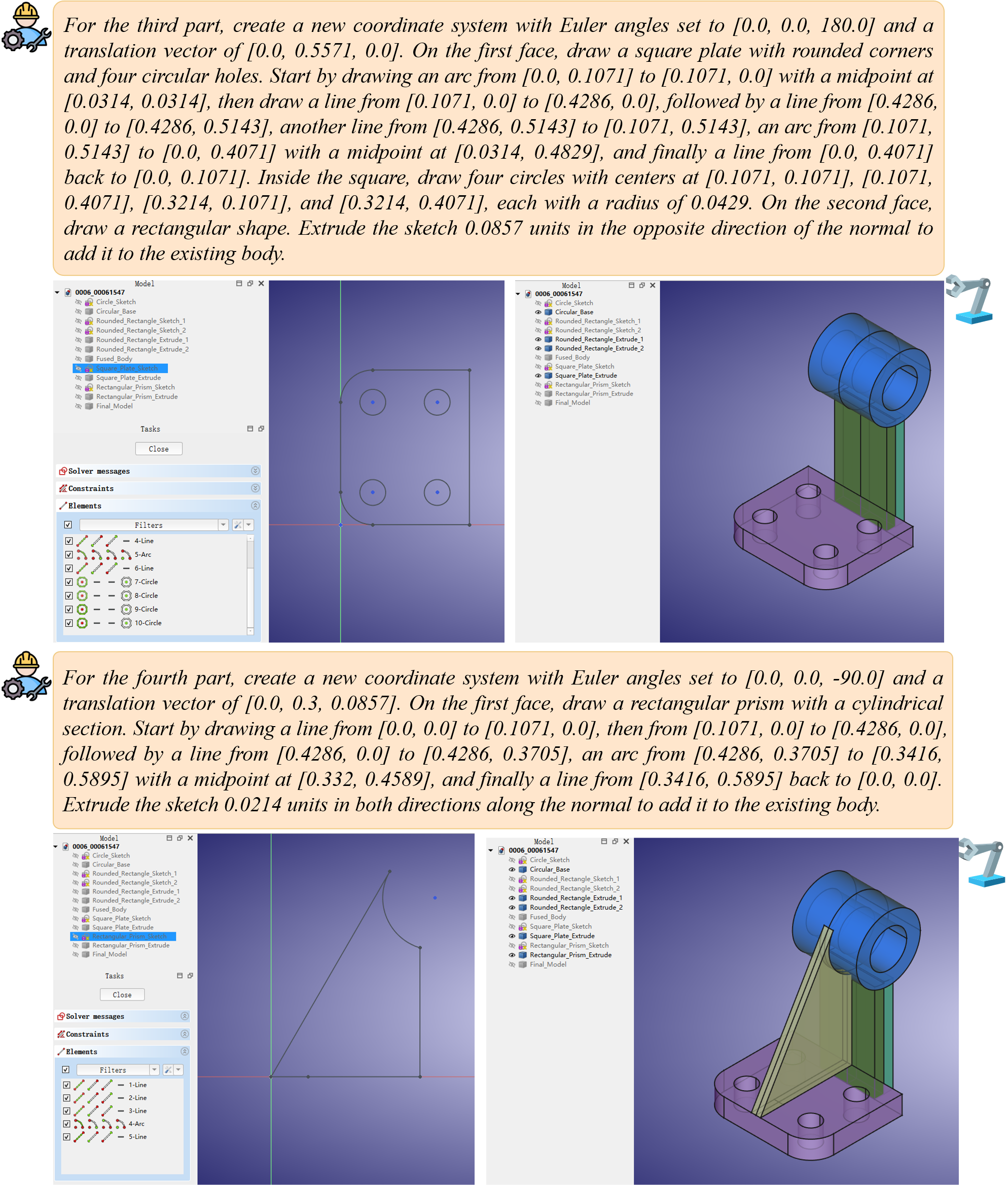}
\end{figure*}

\begin{figure*}[ht]
\caption{feedback of create\_complex\_sketch}
\label{fig:create_complex_sketch}
\begin{lstlisting}[style=pythoncode]
def create_complex_sketch(elements: List[SketchElement] = None, sketch_name: str = None) -> InterfaceResult:
    
    """
    Creates a composite sketch consisting of multiple geometric elements.  
    This function is used to batch-draw sketch profiles in FreeCAD composed of lines, circles, arcs, or splines.  
    Args:
        elements (List[SketchElement]): A list of geometric elements
        sketch_name (str): Name of the sketch.
    """
    # Call FreeCAD API to draw elements
     ...
    # Return interface result
    # 1. Successfully created sketch and derived face.
    return InterfaceResult(True,[Action(description='Successfully created sketch'+sketch_name+'and its sketch-derived face'+face_name])
    # 2. Sketch creation failed due to geometric issues, suggest step-by-step creation 
    return InterfaceResult(False,[Action(description='Sketch creation failed: '+ feedback+'. Please try creating each profile loop one by one.')])
    # 3. Execution error occurred during sketch generation (include exception message)
    return InterfaceResult(False,[Action(description='Sketch creation failed due to an internal error:'+str(error_message))])
\end{lstlisting}
\end{figure*}

\begin{figure*}[ht]
\caption{feedback of bool$\_$operation}
\label{fig:bool}
\begin{lstlisting}[style=pythoncode]
def boolean_operation(
    base_object_name: str,
    tool_object_name: str,
    operation: Literal["cut", "fuse", "common"],
    name: str = None,
) -> InterfaceResult:
    """
    Performs a boolean operation (cut, fuse, or common) between two solid objects to create a new 3D model entity.
    
    This function executes the specified boolean operation between the base object
    and the tool object based on the given operation type:
    - "cut": subtracts the tool object from the base object;
    - "fuse": merges the base and tool objects;
    - "common": computes the intersection of the base and tool objects.
    
    Args:
        base_object_name (str): The name of the base object involved in the boolean operation.
        tool_object_name (str): The name of the tool object used in the boolean operation.
        operation (Literal["cut", "fuse", "common"]): The type of boolean operation to perform.
        name (str): The name of the resulting object. 
    """
    # Call FreeCAD API to perform Boolean operations.
     ...
    # Return interface result
    # 1. Successfully created a new solid as a result of the Boolean operation.
    return InterfaceResult(True, [Action(description='A new solid'+name+' was created by performing the Boolean operation'+operation'.'])
    # 2. Boolean operation failed; include operation type, object names, and error message. 
    return InterfaceResult(False,[Action(description='The Boolean operation '+operation+' between base object '+base_object_name+' and tool object '+tool_object_name+' failed. Error: {str(e)}")
])
\end{lstlisting}
\end{figure*}

\begin{algorithm}[ht]
\caption{Online Curriculum RL for CAD Tool-Using Agents}
\label{alg:online_curriculum_rl}
\KwIn{
    $\mathcal{T}_{\text{pool}}$: Held-out CAD task pool\;
    $\pi_{\theta}$: Policy initialized via $\mathcal{M_{SFT}}$\;
    $\mathcal{E}$: CAD environment\;
    $\mathcal{V}$: Held-in test set\;
    $\alpha$: curriculum decay factor $(0{<}\alpha{<}1)$\;
    $N_{\text{update}}$: policy update interval;
}
\KwOut{Optimized policy $\pi_{\theta}$}

Initialize replay buffer $\mathcal{B} \leftarrow \emptyset$\;
Group tasks by part-count levels: $\{\mathcal{T}_1, ..., \mathcal{T}_L\}$\;
Group held-in validation set similarly: $\{\mathcal{V}_1, ..., \mathcal{V}_L\}$\;
Set curriculum level $\ell \leftarrow 1$\;
Initialize running window of recent perplexities $\mathcal{P}_{\text{recent}} \leftarrow []$\;

\While{$\ell \leq L$}{
    Compute initial perplexity threshold for current level:\;
    \[
    \delta_\ell' = \frac{1}{|\mathcal{V}_\ell|} \sum_{\tau \in \mathcal{V}_\ell}
    \exp\left(-\frac{1}{|\tau|} \sum_{k=1}^{|\tau|} \log \pi_\theta(a_k | s_k) \right)
    \]
    Set $\delta \leftarrow \alpha \cdot \delta_\ell'$\;

    \While{mean$(\mathcal{P}_{\text{recent}}) \geq \delta$}{
        Sample task $t \sim \mathcal{T}_{\ell}$\;
        Rollout trajectory $\tau = \{(s_k, a_k, r_k)\} \leftarrow \mathcal{E}(\pi_{\theta}, t)$\;
        Store trajectory in buffer: $\mathcal{B} \leftarrow \mathcal{B} \cup \{\tau\}$\;

        Compute perplexity for $\tau$: \\
        $P(\tau) = \exp\left(-\frac{1}{|\tau|} \sum_{k=1}^{|\tau|} \log \pi_\theta(a_k | s_k) \right)$\;
        Update recent perplexities: $\mathcal{P}_{\text{recent}} \leftarrow \mathcal{P}_{\text{recent}} \cup \{P(\tau)\}$\;

        \If{$|\mathcal{B}| \geq N_{\text{update}}$}{
            Update policy $\pi_\theta$ using GRPO on $\mathcal{B}$\;
        }
    }

    Advance to next curriculum level: $\ell \leftarrow \ell + 1$\;
    Reset $\mathcal{P}_{\text{recent}} \leftarrow []$\;
}
\Return{$\pi_{\theta}$}
\end{algorithm}
\end{document}